\documentclass[letterpaper]{article} 
\usepackage{aaai25}  
\usepackage{times}  
\usepackage{helvet}  
\usepackage{courier}  
\usepackage[hyphens]{url}  
\usepackage{graphicx} 
\usepackage{natbib}  
\usepackage{caption} 
\usepackage{algorithm}
\usepackage{algorithmic}

\usepackage{newfloat}
\usepackage{listings}
\DeclareCaptionStyle{ruled}{labelfont=normalfont,labelsep=colon,strut=off} 
\floatstyle{ruled}
\newfloat{listing}{tb}{lst}{}
\floatname{listing}{Listing}
\usepackage{graphicx,subfigure}
\usepackage{amsmath}

\DeclareMathOperator*{\argmin}{argmin}
\usepackage{amssymb}
\usepackage{algorithm}
\usepackage{multirow}
\usepackage{svg}
\usepackage{enumerate}
\usepackage{makecell}
\usepackage{enumitem}
\usepackage[symbol]{footmisc}

\begin{document}
%
\title{SAP: Corrective Machine Unlearning with {\em S}caled {\em A}ctivation {\em P}rojection  for Label Noise Robustness }
\author {
    Sangamesh Kodge,
    Deepak Ravikumar,
    Gobinda Saha,
    Kaushik Roy\\
    }
\affiliations{
    Elmore Family School of Electrical and Computer Engineering\\
    Purdue University, West Lafayette, Indiana, USA\\
    skodge@purdue.edu, dravikum@purdue.edu, gsaha@purdue.edu, kaushik@purdue.edu
}

\maketitle
\begin{abstract}
Label corruption, where training samples are mislabeled due to non-expert annotation or adversarial attacks, significantly degrades model performance. 
Acquiring large, perfectly labeled datasets is costly, and retraining models from scratch is computationally expensive.
To address this, we introduce {\em S}caled {\em A}ctivation {\em P}rojection ({\em SAP}), a novel SVD (Singular Value Decomposition)-based corrective machine unlearning algorithm.
{\em SAP} mitigates label noise by identifying a small subset of trusted samples using cross-entropy loss and projecting model weights onto a clean activation space estimated using SVD on these trusted samples. 
This process suppresses the noise introduced in activations due to the mislabeled samples. 
In our experiments, we demonstrate {\em SAP}'s effectiveness on synthetic noise with different settings and real-world label noise. 
{\em SAP} applied to the CIFAR dataset with $25\%$ synthetic corruption show upto $6\%$ generalization improvements. Additionally, {\em SAP} can improve the generalization over noise robust training approaches on CIFAR dataset by $\sim3.2\%$ on average.
Further, we observe generalization improvements of $2.31\%$ for a Vision Transformer model trained on naturally corrupted Clothing1M. 
\end{abstract}
%
 \begin{links}
 \link{Code}{https://github.com/sangamesh-kodge/SAP.git}
 \end{links}
\section{Introduction}  \label{sec:intro}
\begin{figure}[t]
    \centering
    \includegraphics[width=0.90\columnwidth]{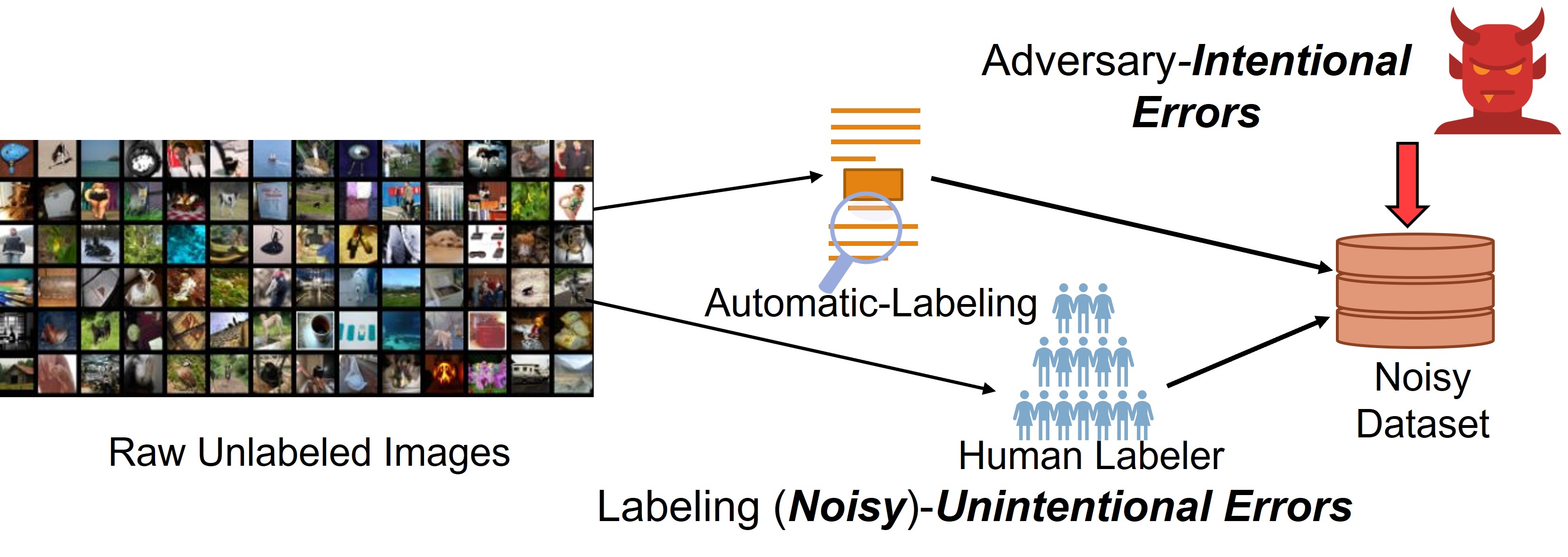}
    \caption{Sources of label noise - Intentional/Unintentional Errors.}
    \label{fig:dataset_creation}
\end{figure}

Deep learning models have revolutionized various fields like natural language processing and computer vision, largely due to the availability of massive datasets. 
However, the very scale of these datasets presents a challenge: guaranteeing accurate labeling as shown in Figure~\ref{fig:dataset_creation}. 
Recent studies \cite{northcutt2021pervasive}, have exposed the significant presence of labeling errors in widely used benchmarks.
These errors can be unintentional, arising from human error \cite{sambasivan2021everyone} or in recent times automated labeling using Large Language models \cite{wang2024human,pangakis2024knowledgedistillationautomatedannotation}. 
Alternatively, they can be deliberate, introduced through attacks like data poisoning \cite{schwarzschild2021just,biggo2012poisoning,jagielski2018manipulating,chen2017targeted,chen2022amplifying}. 
Regardless of origin, these errors degrade model performance.
While deep learning models exhibit robustness to a certain degree of label noise, they are not immune to label noise, as demonstrated in \cite{steinhardt2017certified}.

Corrective Machine Unlearning \cite{goel2024correctivemachineunlearning} is emerging as a promising solution to the aforementioned challenge, with the objective of reducing the impact of mislabeled data by manipulating the trained model.
A vast majority of unlearning algorithms \cite{triantafillou2024makingprogressunlearningfindings} require knowledge of which samples are mislabeled to partition the data into the retain set and the forget set.
It is challenging to distinguish between mislabeled samples and hard to learn samples \cite{garg2023memorization}. This presents a major challenge to deploy machine unlearning algorithms for tackling label noise.

This work introduces {\em SAP}, a novel algorithm based on Singular Value Decomposition (SVD) to improve model generalization in the presence of label noise. Notably, our approach achieves this improvement with a single update to the model weights.
{\em SAP} leverages a small set of samples identified as having low cross-entropy loss. These samples are hypothesized to be correctly labeled and form a ``clean" subset.
We utilize this clean subset and SVD to compute an update for the model weights. 
This update involves estimating a clean activation space and projecting the model weights onto this space. 
This projection aims to suppress activations corresponding to potentially corrupted activations, improving model performance on unseen data.
As {\em SAP} relies on the estimated clean subset (or retain set), we avoid explicitly detecting mislabeled data, which can be more difficult to obtain. 

In order to visualize the effectiveness of {\em SAP}, we trained a network for binary classification task on clean training data (clean model) and corrupt data (corrupted model), where $10\%$ labels were flipped from 2D spiral dataset \cite{manifoldmixup}.
We present a detailed experimental details for this toy illustration in the Supplementary. 
Figure~\ref{fig:boundary_clean_dataset} and \ref{fig:boundary_corrupt_dataset} show the decision boundary of the model for the former and later cases respectively. 
We observe that, with label corruption, the test accuracy of the model drops by $4.8\%$ from the model trained without any label corruption. 
The model trained on noisy data  in Figure~\ref{fig:boundary_corrupt_dataset}  has a more complex and rough decision boundary which indicates memorization of the corrupt data points leading to a loss in generalization. 
When {\em SAP} is applied to the model trained on corrupt data as seen in Figure~\ref{fig:boundary_sap}, we observe improved decision boundary smoothness and  recover the generalization performance as hypothesized.
\begin{figure}[t]
\centering
    \subfigure[Clean model]{
        \label{fig:boundary_clean_dataset}
        \includegraphics[width=0.30\columnwidth]{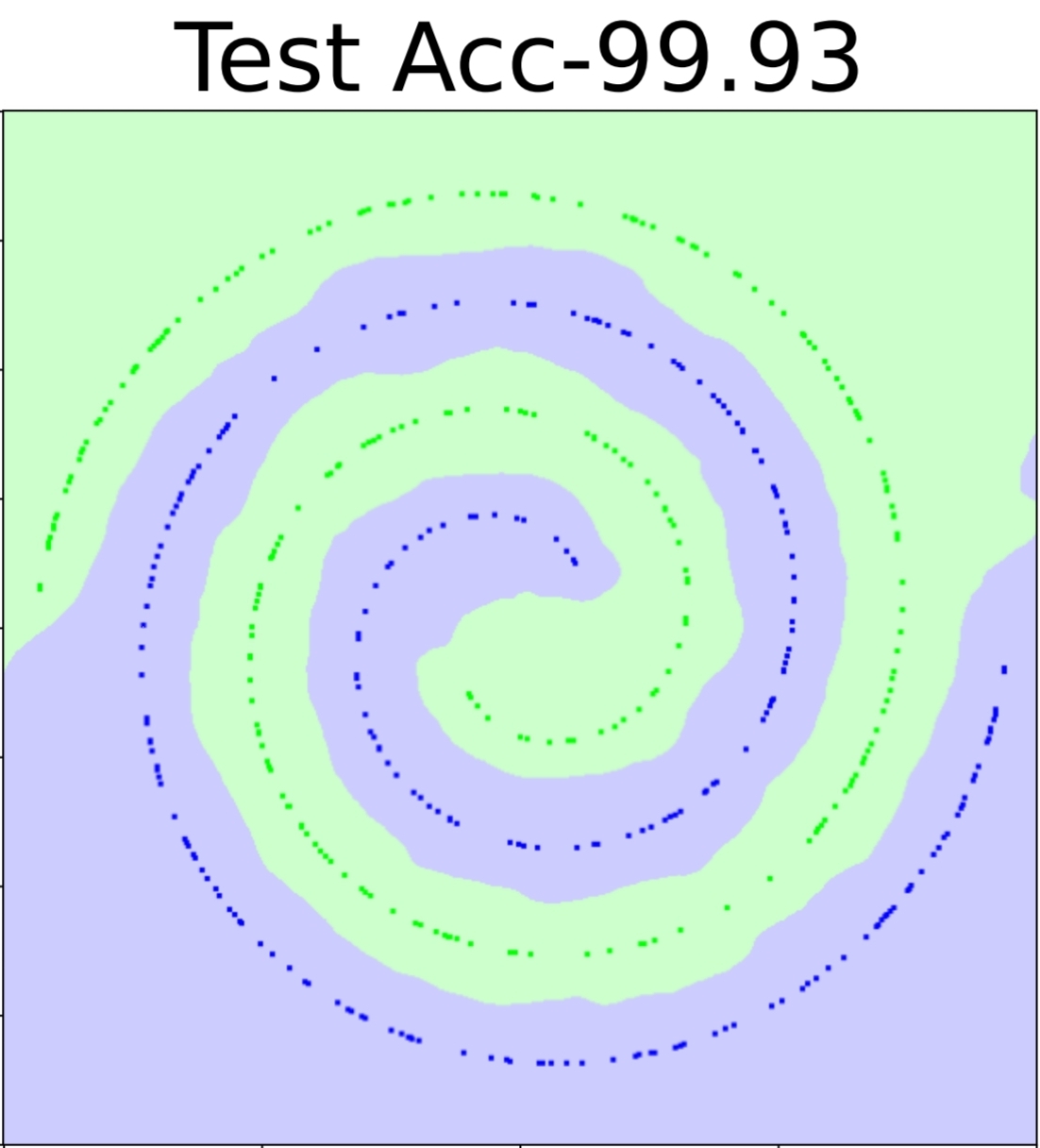}
    }      
    \subfigure[Corrupt model]{
        \label{fig:boundary_corrupt_dataset}
        \includegraphics[ width=0.30\columnwidth]{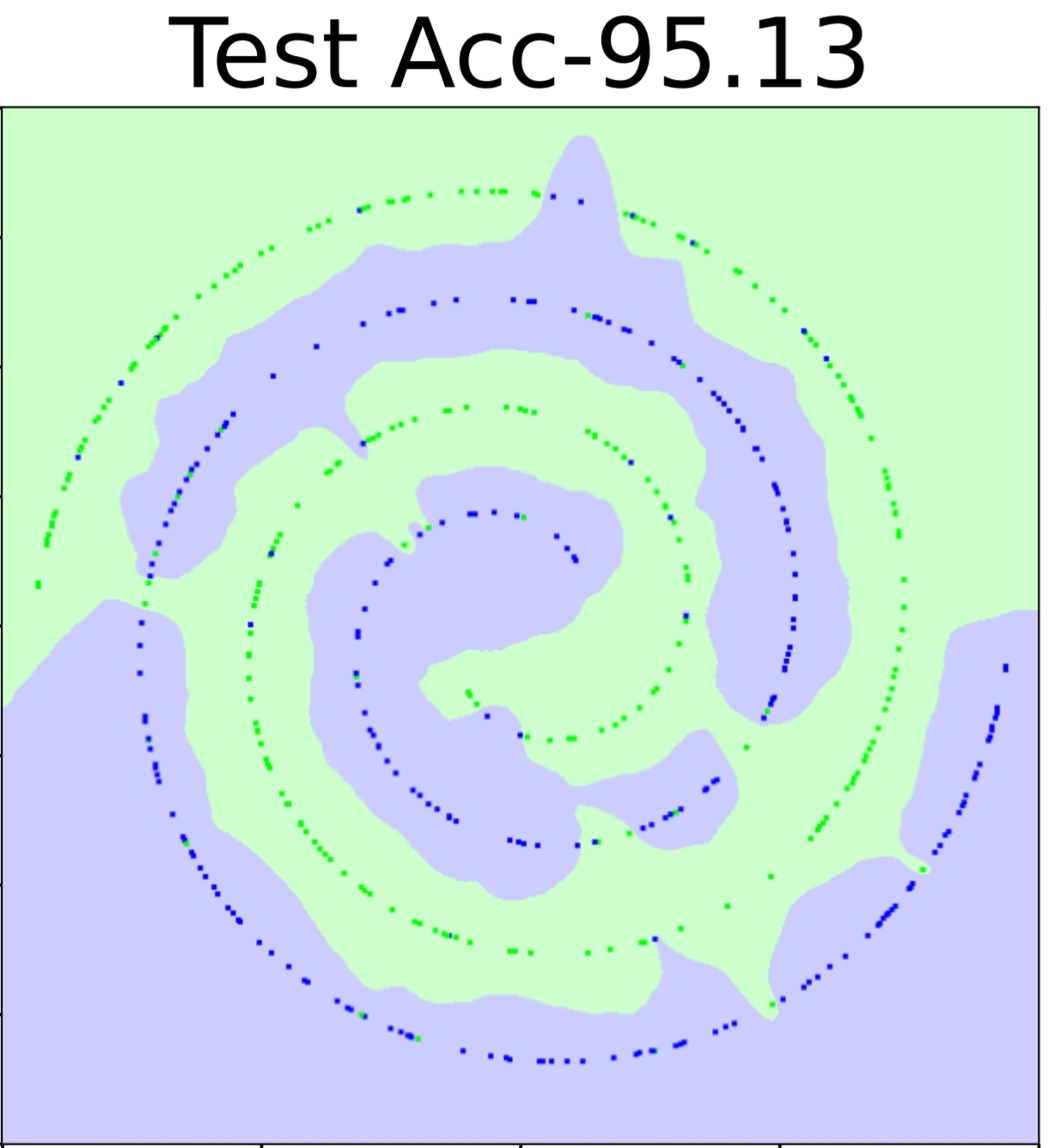}
    }    
    \subfigure[SAP applied to corrupt model]{
        \label{fig:boundary_sap}
        \includegraphics[ width=0.30\columnwidth]{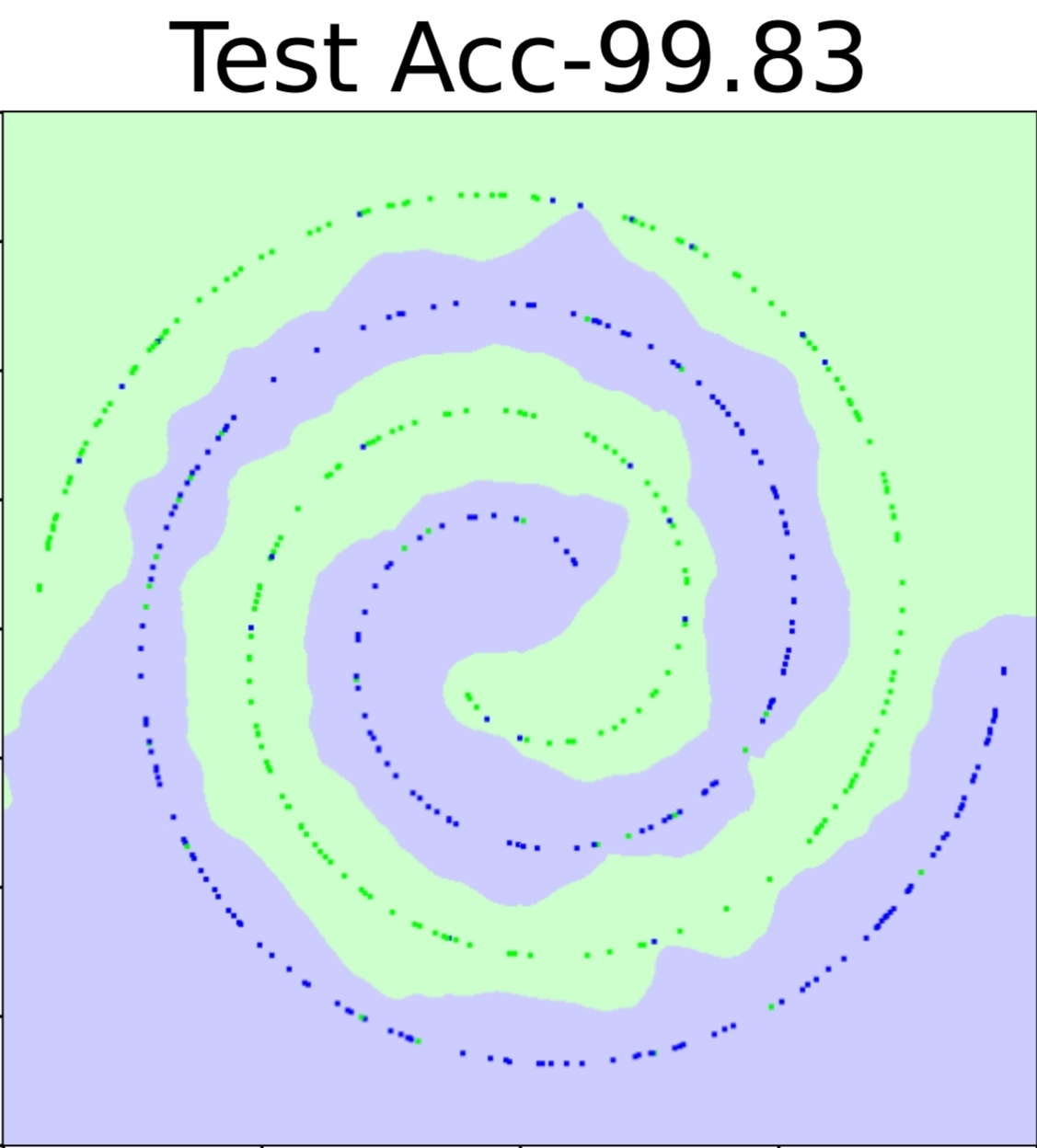}
    }
\caption{Decision boundaries for a network trained the 2D spiral data (a) with full Clean data, (b) with $10\%$ corrupt data and (c) {\em SAP} applied to network trained on corrupt data.}
\label{fig:boundary}
\vspace{-2mm}
\end{figure}
We summarize the contributions of our work below:
\begin{itemize}[noitemsep,topsep=0pt]
    \item We propose {\em SAP}, a novel corrective unlearning algorithm based on Singular Value Decomposition (SVD). {\em SAP} automates the selection of clean samples using the model loss and updates model weights in a single step, leading to computational efficiency. Notably, we alleviate the challenge of explicitly detecting mislabeled data which is required by most of the unlearning algorithms.
    \item We empirically validate {\em SAP}'s performance on synthetic and real-world label noise scenarios across various model architectures and datasets, demonstrating generalization improvements of up to $6\%$ and $2.31\%$, respectively.
\end{itemize}

\section{Background}
\label{sec:background}
\subsection{Singular Value Decomposition}
A rectangular matrix $A$ in $\mathbb{R}^{d\times n}$ can be decomposed with Singular Value Decomposition (SVD) \cite{Deisenroth2020} as 
\begin{equation}\label{eqn:svd}
    A = U \Sigma V^T.
\end{equation}
Here, $U\in\mathbb{R}^{d\times d}$ and $V \in\mathbb{R}^{n\times n}$ are orthogonal matrices and are called left singular matrix and right singular matrix, respectively.
$\Sigma \in \mathbb{R}^{d\times n}$ is a diagonal matrix comprising singular values. Each column vector of $U$ is $d$-dimensional and these vectors form the basis of the column space of $A$. 
The amount of variance explained by the $i^{th}$ vector of $U$, $u_i$, having singular value of $\sigma_i$ is proportional to $\sigma_i^2$ and hence the fraction variance explained is given by 
\begin{equation}
    \label{eqn:normalized_sigma}    
    \widetilde{\sigma}_i = \sigma_i^2/ (\sum_{j=1}^d (\sigma_j^2)).
\end{equation}

\subsection{Related Work}
Corrective Machine Unlearning \cite{goel2024correctivemachineunlearning} has emerged as a promising approach to tackle the challenge of mitigating the impact of corrupted data, such as mislabeled samples, on trained models.
Unlike traditional machine unlearning \cite{triantafillou2024makingprogressunlearningfindings},  which often focuses on privacy concerns when removing data, corrective unlearning prioritizes improving model generalization, bypassing the need for adhering to specific privacy requirements \cite{hayes2024inexactunlearningneedscareful}.
In the context of label noise, corrective unlearning aims to improve the model's generalization on unseen data, when the training dataset contains mislabeled samples. 
Recent work by \cite{goel2024correctivemachineunlearning} benchmarks several state-of-the-art (SoTA) unlearning algorithms, such as SSD \cite{ssd}, CF-k \cite{cfk} and SCRUB \cite{scrub}, within the corrective unlearning framework. 
Notably, ASSD \cite{assd} proposes an extension of SSD which automatically chooses the suitable unlearning hyperparameters to handle the label errors on supply chain delay prediction problem.

Additionally, model generalization can be further improved by using these algorithms in conjunction with traditional approaches such as data filtering \cite{northcutt2021confident,jia2022learningtrainingdynamicsidentifying,maini2022characterizing}, sample selection \cite{cheng2021learninginstancedependentlabelnoise}, label correction \cite{zheng2021meta}, regularization \cite{wei2021smooth,wei2023mitigatingmemorizationnoisylabels}, robust loss functions \cite{wang2019symmetric}, optimizations \cite{foret2021sharpnessaware}, curriculum learning \cite{jiang2018mentornet,zhang2018generalized}, data augmentation \cite{mixup,mentormix}, and noise transition matrix estimation \cite{pmlr-v162-zhu22k,Cheng_2022_CVPR}. We provide a brief Literature Survey on Label Noise Learning in the supplementary.

\section{{\em S}caled {\em A}ctivation {\em P}rojection Algorithm}
\label{sec:method}
\begin{figure}[b]
    \centering
    \vspace{-3mm}
    \includegraphics[width=1.0\columnwidth]{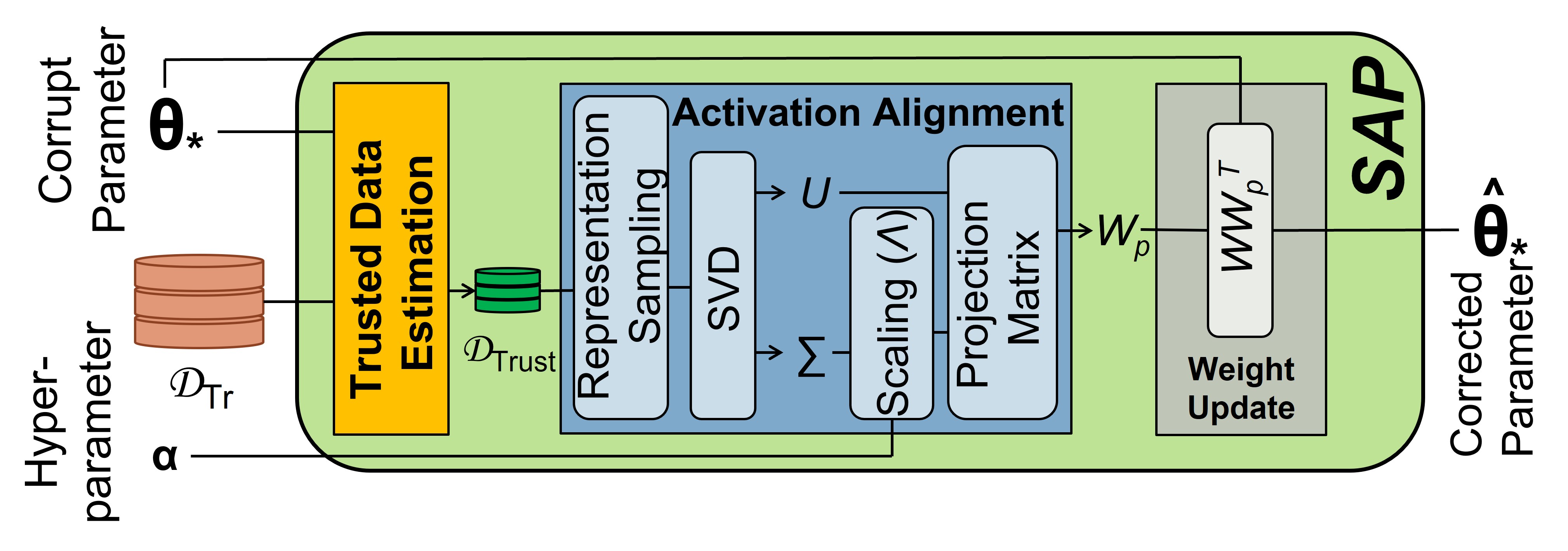}
    \caption{Overview of the proposed {\em SAP} algorithm. Here, $\theta_*$ denotes the model weights which comprises of the layer weights represented by $W$. }
    \label{fig:overview}
\end{figure}
A network trained on the training dataset, $\mathcal{D}_{Tr}$, with corrupted labels is prone to overfit on the incorrect samples within the training data, as seen in Figure~\ref{fig:boundary_corrupt_dataset}.
Such a network is likely to produce spurious intermediate activations to incorporate the incorrect sample-label mapping. 
{\em SAP} (Figure~\ref{fig:overview}), outlined in Algorithm~\ref{alg:our}, aims to align these intermediate activations with representative or clean activations.
In the next Subsection, we explain our weight update mechanism which enables us to modify the model weights to remove the influence of corrupt data.
We provide our codebase at \url{https://github.com/sangamesh-kodge/SAP.git}.

\subsection{Weight Update}\label{subsec:weight_update}
\begin{figure*}[t]
\centering
    \subfigure[Singular Values]{
        \label{fig:singular_value}
        \includegraphics[trim={0.2cm 0cm 5cm 2cm},clip,width=0.18\textwidth]{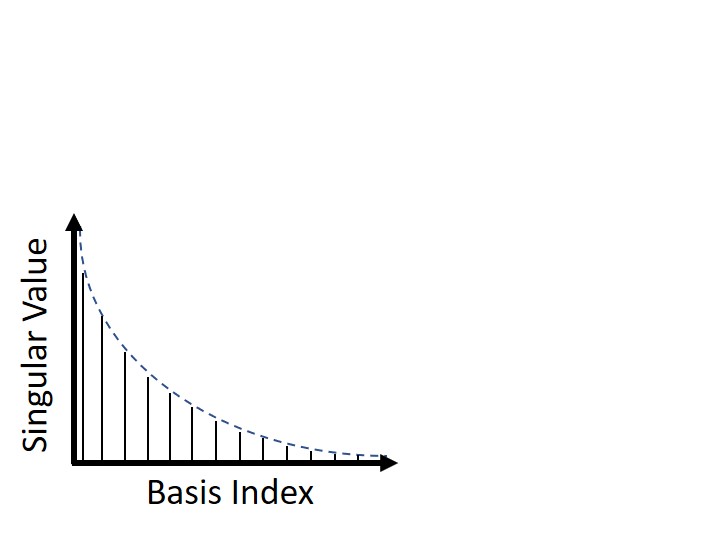}
    }
    \quad
    \subfigure[Importance Scaling]{
        \label{fig:importance}
        \includegraphics[trim={0.2cm 0cm 3cm 2cm},clip,width=0.25\textwidth]{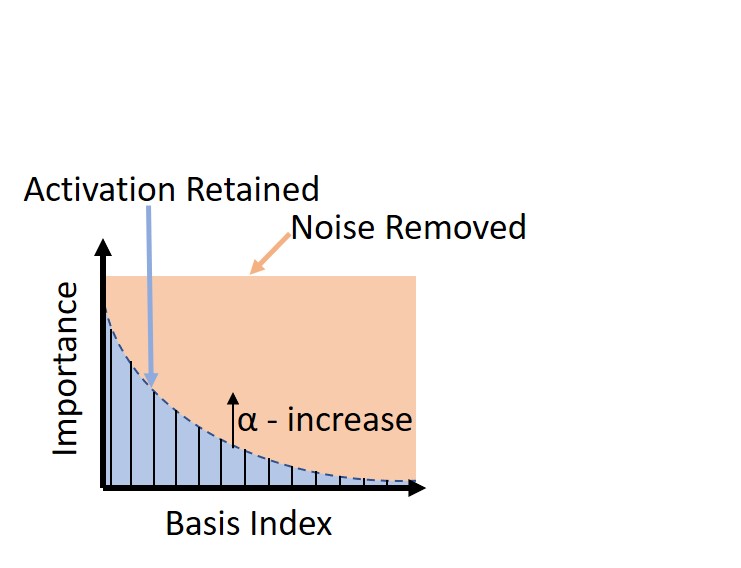}
    }
    \quad
    \subfigure[Activation Alignment]{
        \label{fig:projection}
        \includegraphics[width=0.35\textwidth]{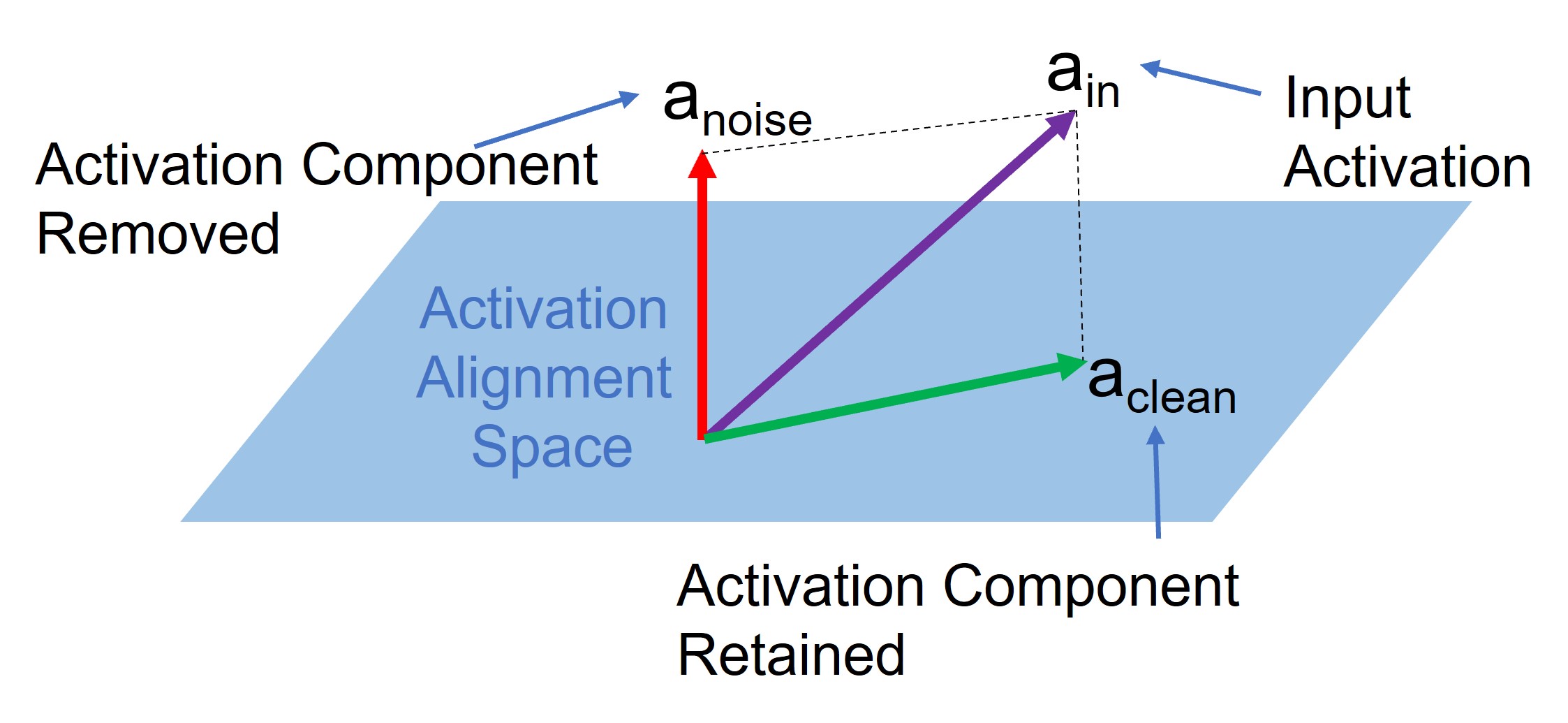}
    }
\caption{(a) Singular value distribution obtained through SVD on the Representative activations. (b) Effect of $\alpha$ in importance scaling Equation~\ref{eqn:scaling} and (c) the projection of noisy activation on Activation Alignment Space to obtain clean activations. }
\label{fig:scaling}
\vspace{-3mm}
\end{figure*}

Consider a linear layer with parameters $W$ that generates the output activation $a_{out}$, given by $a_{out} = a_{in} W^T$.
We align the input activations of this layer with the trusted activations by projecting the activations $a_{in}$ onto a strategically constructed projection matrix $W_p$, also known as the Activation Alignment matrix.
This projection would suppress the noisy activation as shown in Figure~\ref{fig:projection}.
This results in updating the output activation as $\widehat{a}_{out} = (a_{in} W_p) W^T$. 
We rewrite this equation to absorb the alignment matrix $W_p$ into the layer weights, effectively updating the weights as $\widehat{W} = W W_p^T$, as detailed in Equation~\ref{eqn:weight_update}. 
This weight update rule explains the \texttt{update\_parameter} procedure in line 7 of Algorithm~\ref{alg:our}. 
In the rest of this Section, our focus is on obtaining the alignment matrix $W_p$.
\begin{equation}
    \label{eqn:weight_update}
    \widehat{a}_{out} = \underbrace{(a_{in} W_p)}_\text{Activation Projection} \hspace{-4mm} W^T =  a_{in}  (\hspace{-2mm}\underbrace{W W_p^T}_\text{Weight Update}\hspace{-2mm})^T = a_{in} \widehat{W}^T
\end{equation}

\begin{algorithm}[t]
\textbf{Input:} $\theta_*$ is the parameters of the original model; $\mathcal{D}_{Tr}$ is the training dataset with corrupt labels, and $\alpha$ is a hyperparameter called scaling coefficient.

\textbf{procedure} \texttt{SAP}( $\theta_*$, $\mathcal{D}_{Tr}$, $\alpha$)\\
1. $\mathcal{D}_{Trust}$ = \texttt{get\_trusted}($\theta_*$, $\mathcal{D}_{Tr}$)\hfill\textcolor{gray}{// Eqn~\ref{eqn:trusted_data}} \\  
2. $R$ = \texttt{representation}($\theta_*$, $\mathcal{D}_{Trust}$) \hfill\textcolor{gray}{// Eqn.~\ref{eqn:linear_representation}, \ref{eqn:conv_representation} } \\ 
3. \textbf{for} each layer with parameter $W^l_*$ $\in \theta_*$  \textbf{do}\\
4. \hspace{4mm}$U^l$, $\Sigma^l$ = \texttt{SVD}($R^l$) \hfill\textcolor{gray}{//  Eqn.~\ref{eqn:svd}} \\ 
5. \hspace{4mm}$\Lambda^l$ = \texttt{scale\_importance}($\Sigma^l$, $\alpha$ )\hfill\textcolor{gray}{// Eqn.~\ref{eqn:scaling}} \\
6. \hspace{4mm}$W_p^l$ = $U^l  \Lambda^l  (U^l)^T$\hfill\textcolor{gray}{// Eqn.~\ref{eqn:projection}} \\
7. \hspace{4mm}$\widehat{W}^l_* =$ \texttt{update\_parameter}($W_p^l$, $W^l_*$ )\hfill\textcolor{gray}{// Eqn.~\ref{eqn:weight_update}} \\
8. \textbf{return} $\hat{\theta}_*$ \hfill\textcolor{gray}{// where $\hat{\theta}_*$ is updated parameters}
\caption{ SAP algorithm. }
\label{alg:our}
\end{algorithm}

Obtaining a high-quality alignment matrix requires access to representative activations for `trusted' samples or the samples in $\mathcal{D}_{Tr}$ with correct labels.
We, therefore, divide the step of obtaining $W_p$ into two parts: (1) approximately estimating a few correctly labeled or `trusted' samples from $\mathcal{D}_{Tr}$, and (2) obtaining the Activation Alignment Matrix using these samples.

\subsection{Trusted Data Estimation}\label{subsec:trusted_data}
We define the trusted dataset, $\mathcal{D}_{Trust}=\{(x_i, y_i)\}_{i=1}^{N_{Trust}}$, as a subset of $\mathcal{D}_{Tr}$ containing a few ($N_{Trust}$) correctly labeled samples. 
This work proposes a method to automate the process of identifying these trusted samples.
We leverage the cross-entropy loss, denoted by $\mathcal{L}$, of the trained model with parameters $\theta_*$.
The rationale is that samples with lower cross-entropy loss are more likely to be correctly labeled.
We select $N_{Trust}$ samples with the lowest cross-entropy loss values to from the trusted dataset $\mathcal{D}_{Trust}$. This process can be mathematically expressed by Equation~\ref{eqn:trusted_data} corresponds to \texttt{get\_trusted} of Algorithm~\ref{alg:our}. 
Note, these trusted samples can also be obtained by employing human experts, however, this might be tedious or impractical for larger dataset.  
\begin{align}\label{eqn:trusted_data}
    \begin{split}
        \mathcal{D}_{Trust} &= \argmin_\mathcal{S} \sum_{(x_i, y_i) \in S }(\mathcal{L}(\theta_*, x_i, y_i)); \\
        &\text{such that  } \mathcal{S} := \{\mathcal{S}_i \subseteq  \mathcal{D}_{Tr} \hspace{2mm}|\hspace{2mm}|\mathcal{S}_i| = N_{Trust} \} 
    \end{split}
\end{align}

\subsection{Activation Alignment}\label{subsec:activation_alignment}
To estimate the activation basis for the clean samples, we gather the input activations of linear and convolutional layers, as detailed below.

\subsubsection{Representation Sampling:}\label{subsec:representation}
For the $l^{th}$ linear or convolutional layer of the network, we collect the input activation $a_{in}$ and store these activation in a representation matrix $R^l$. 
This matrix captures representative information from the trusted samples in the input activations. 
It is utilized to estimate the trusted activation basis.
Next, we elaborate on the details of this representation matrix~\cite{saha2021gradient} for each type of layer.

\begin{enumerate}
    \item  \textit{ Linear Layer -} For a linear layer, the representation matrix, as given by Equation~\ref{eqn:linear_representation}, stores the input activations for all the samples in $\mathcal{D}_{Trust}$.
    \begin{equation}
    \label{eqn:linear_representation}
        R_{linear} = [(a_{in}^i)_{i=1}^{N_{Trust}}]   
    \end{equation}
    Here, $a_{in}$ is the input activation for the linear layer. 
    
    \item  \textit{ Convolutional Layer -} 
    For the convolutional layer, we express the convolution operation as matrix multiplication to apply the weight update rule proposed in Equation~\ref{eqn:weight_update}. 
    This is achieved through the \texttt{unfold} \cite{chetlur2014cudnnefficientprimitivesdeep} operation where the convolution operation is represented as matrix multiplication. 
    Let the kernel size of the convolutional layer be $C_{out} \times C_{in} \times k \times k $, where $C_{in}$ represents the number of input channels, $C_{out}$ denotes the number of output channels and $k$ is the kernel size.
    The unfold operation would flatten all the patches in the input activations on which this kernel operates in a sliding window fashion, which gives us a matrix of size $n_p\times C_{in}kk$. 
    Here, $n_p$ represents the number of patches in the activation $a_{in}^i$ of the $i^{th}$ sample in $\mathcal{D}_{Trust}$. 
    This process allows us to represent convolution as matrix multiplication between the 
    reshaped weights of size $C_{out}\times{C_{in}kk}$ and the unfolded activation of $a^i_{in}$ of size $n_p\times{C_{in}kk}$.
    Figure~8 in supplementary details the conversion of the convolution operation into the matrix multiplication operation.
    The unfolded activations of all the samples are concatenated and stored in the representation matrix as represented by Equation~\ref{eqn:conv_representation}. 
    We use the reshaped convolutions parameters to update the parameters as given in Equation~\ref{eqn:weight_update}. 
    \begin{equation}
    \label{eqn:conv_representation}
        R_{conv} = [(\texttt{unfold}(a_{in}^i)^T)_{i=1}^{N_{Trust}}]   
    \end{equation}
    
    The \texttt{representation} procedure in line 2 of Algorithm~\ref{alg:our} obtains a list of these representation matrices $R$. 
    The $l^{th}$ element of this list, $R^l$, is the representation matrix of the $l^{th}$ linear or convolutional layer of the network.
\end{enumerate}

\subsubsection{SVD on Representations:} 
We perform Singular Value Decomposition (SVD), as given in Equation~\ref{eqn:svd}, on the representation matrices $R^l$ for the $l^{th}$ layer, as shown in line 4 of Algorithm~\ref{alg:our}.
The \texttt{SVD} procedure provides us with the basis vectors $U$ and the singular values $\Sigma$ for the trusted samples.
The $j^{th}$ vector in $U$, $u_j$, is a vector having a unit norm and does not capture the importance of this vector.
We scale these vectors in proportion to the variance explained by these vectors.

\subsubsection{Importance Scaling:} We formulate a diagonal importance matrix $\Lambda$ with the $j^{th}$ diagonal component $\lambda_j$ given by Equation~\ref{eqn:scaling} \cite{kodge2024deep}. 
Here, $\widetilde{\sigma}_j$ represents the normalized $j^{th}$ singular value, as defined in Equation~\ref{eqn:normalized_sigma}. 
The effect of such scaling on singular values is shown in Figure~\ref{fig:scaling}.
The parameter $\alpha \in (0, \infty)$, known as the scaling coefficient, serves as a hyperparameter controlling the scaling of the basis vectors. 
When $\alpha$ is set to 1, the basis vectors are scaled by the amount of variance they explain. 
Figure~\ref{fig:importance} shows the effect of changing $\alpha$ on the scaling of the basis vectors.
As $\alpha$ increases, the importance score for each basis vector increases and approaches 1 as $\alpha \to \infty$. Conversely, a decrease in $\alpha$ diminishes the importance of the basis vectors, approaching 0 as $\alpha \to 0$.
The \texttt{scale\_parameter} procedure in line 5 of Algorithm~\ref{alg:our} corresponds to this importance-based scaling. 

\begin{equation}
    \label{eqn:scaling}
    \lambda_i  = \frac{\alpha  \widetilde{\sigma}_i}{(\alpha - 1 )  \widetilde{\sigma}_i + 1}
\end{equation}

\subsubsection{Projection Matrix:} The Activation Alignment Matrix $W_p$ is obtained in line 6 of Algorithm~\ref{alg:our} using the basis vectors $U$ and the scaled importance matrix $\Lambda$ by Equation~\ref{eqn:projection}.
This matrix projects the input activation into the activations spanned by the trusted samples effectively removing the noise due to the corrupted labels from the training dataset. 
\begin{equation}
    \label{eqn:projection}
    W_p = U  \Lambda  (U)^T
\end{equation}

\section{Experiments}
\label{sec:results}
\subsection{Experimental Setup}
We evaluate our algorithm on the CIFAR10 and CIFAR100 datasets \cite{cifar} using VGG11 \cite{vgg} and ResNet18 \cite{he2016deep} models.
To simulate various noise scenarios, we introduce synthetic noise to the original training data according to Equations~\ref{eqn:sym_noise},~\ref{eqn:asym_noise}, or ~\ref{eqn:real_noise} and subsequently partition the data into a $95\%$ training set ($\mathcal{D}_{Tr}$) and a $5\%$ validation set ($\mathcal{D}_{Val}$). Importantly, we add noise before the train-validation split, mimicking a real-world scenario where noise is present from the outset. Unless otherwise specified, we tune hyperparameters on validation set ($\mathcal{D}_{Val}$) and  use the accuracy on the test set  ($\mathcal{D}_{Te}$) as the metric to assess generalization performance of the algorithm. In all our tables we report the mean and standard deviation across three randomly chosen seeds in all our experiments. Specific training configurations and hyperparameters are provided in the supplementary material for all the experiments. We explore different type of noise in our evaluations presented in the next subsection.   

\begin{table*}[t]
\vspace{-1mm}
\begin{center}
\resizebox{1.0\textwidth}{!}{
    \begin{tabular}{l|l|c|c||cc||cc|cc|cc||c}
    \Xhline{3\arrayrulewidth}
    & \multirow{3}{*}{\bf Method} & & &  \multicolumn{2}{c||}{\bf \texttt{VGG11\_BN}} &  \multicolumn{6}{c||}{\bf \texttt{ResNet18}}&\multirow{3}{*}{\bf Average} \\
            \cline{5-12}
    &&{\bf Retain} &{\bf Forget}  &\multicolumn{2}{c||}{\bf Symmetric Noise ( Eqn.~\ref{eqn:sym_noise} )} &  \multicolumn{2}{c|}{\bf Symmetric Noise ( Eqn.~\ref{eqn:sym_noise} )} &  \multicolumn{2}{c|}{\bf Asymmetric Noise ( Eqn.~\ref{eqn:asym_noise} )} &  \multicolumn{2}{c||}{\bf Hierarchical Noise ( Eqn.~\ref{eqn:real_noise} )}\\
    && {\bf samples}& {\bf samples}& {\bf $\eta=0.1$}& {\bf $\eta=0.25$}& {\bf $\eta=0.1$}& {\bf $\eta=0.25$}& {\bf $\eta=0.1$}& {\bf $\eta=0.25$}& {\bf $\eta=0.1$}& {\bf $\eta=0.25$}\\
        \hline  
        \hline
        \multirow{6}{*}{\bf \rotatebox[origin=c]{90}{CIFAR10}}
            & Retrain  & -& -& $90.18\pm0.14$ & $89.48\pm0.04$
                        & $93.21\pm0.22$ & $92.45\pm0.06$
                        & $93.38\pm0.07$&$92.75\pm0.23$
                        &$93.47\pm0.21$&$93.14\pm0.23$ 
                        & $92.26$\\ 
            \cline{2-13}
            & Vanilla & $0$ &$0$ & $86.04\pm0.17$& $76.68\pm0.48$ 
                        & $88.55\pm0.16$&$79.47\pm0.46$
                        & $88.53\pm0.34$&$79.79\pm1.53$
                        & $91.42\pm0.33$&$86.42\pm0.38$
                        & $84.61$\\
            & Finetune & $5000$ & $0$  &$85.47\pm0.13$ & $80.94\pm0.76$ 
                        & $88.28\pm0.30$ & $85.16\pm0.12$
                        & $87.31\pm0.81$&$82.82\pm0.98$
                        & $91.42\pm0.33$ &$\textbf{88.23}\pm\textbf{0.73}$ 
                        & $86.20$\\
            & SSD   & $5000$ & $1000$    
                        &$86.00\pm0.21$ & $76.77\pm0.58$ 
                        & $88.54\pm0.17$ & $79.48\pm0.50$
                        & $88.52\pm0.35$&$80.36\pm1.45$
                        &$91.42\pm0.33$&$86.39\pm0.42$ 
                        & $84.68$\\
            & SCRUB   & $1000$ & $200$  
                        & $85.88\pm0.35$&$78.90\pm0.25$
                        & $89.50\pm0.17$&$83.77\pm0.44$
                        & $89.50\pm0.22$&$83.60\pm0.14$
                        &$91.65\pm0.12$&$88.00\pm0.58$ 
                        & $86.35$\\
            & SAP    & $0$\textsuperscript{\textdagger}  & $0$   
                        & $\textbf{87.25}\pm\textbf{0.16}$& $\textbf{82.27}\pm\textbf{0.15}$
                        & $\textbf{90.12}\pm\textbf{0.11}$&$\textbf{85.49}\pm\textbf{0.39}$
                        &$\textbf{90.03}\pm\textbf{0.25}$&$\textbf{86.32}\pm\textbf{0.66}$ 
                        &$\textbf{91.87}\pm\textbf{0.22}$&$87.92\pm0.69$ 
                        & $\textbf{87.66}$\\
    \hline
    \hline
        \multirow{6}{*}{\bf \rotatebox[origin=c]{90}{CIFAR100}}
            & Retrain  & - & -  & $65.76\pm0.23$& $63.66\pm0.47$ 
                        & $72.43\pm0.42$&$71\pm0.15$
                        & $73.76\pm0.46$&$71.62\pm0.47$
                        &$72.73\pm0.19$&$71.16\pm0.39$ 
                        & $70.26$\\ 
            \cline{2-13}
            & Vanilla  & $0$ & $0$  & $60.41\pm0.14$& $50.64\pm0.60$ 
                        & $65.84\pm0.33$&$54.75\pm0.45$
                        & $72.98\pm0.13$&$61.86\pm0.60$
                        &$67.45\pm0.12$&$57.82\pm0.21$ 
                        & $61.47$\\ 
            & Finetune& $5000$ & $0$   
                        &$60.26\pm0.11$ & $52.50\pm0.31$ 
                        & $65.97\pm0.35$ & $57.33\pm0.40$ 
                        & $72.98\pm0.13$&$61.29\pm1.13$
                        &$67.55\pm0.15$&$59.98\pm1.11$ 
                        & $62.23$\\
            & SSD  & $5000$ & $1000$      
                        &$60.38\pm0.16$ & $50.62\pm0.60$ 
                        & $65.84\pm0.33$ & $54.77\pm0.42$ 
                        & $72.99\pm0.11$&$61.67\pm0.55$
                        &$67.43\pm0.21$&$57.83\pm0.21$ 
                        &$61.44$ \\
            & SCRUB   & $1000$ & $200$  
                        & $60.93\pm0.09$&$52.11\pm0.63$
                        & $\textbf{67.02}\pm\textbf{0.29}$&$57.36\pm0.40$
                        & $\textbf{73.12}\pm\textbf{0.18}$&$63.37\pm0.69$
                        &$68.20\pm0.13$&$60.24\pm0.33$ 
                        &$62.79$ \\
            & SAP   & $0$\textsuperscript{\textdagger} & $0$ 
                        & $\textbf{61.10}\pm\textbf{0.23}$& $\textbf{53.31}\pm\textbf{0.78}$ 
                        & $66.82\pm0.17$ &$\textbf{58.74}\pm\textbf{0.61}$
                        & $72.92\pm0.30$&$\textbf{63.57}\pm\textbf{0.49}$
                        &$\textbf{68.24}\pm\textbf{0.17}$& $\textbf{60.76}\pm\textbf{0.50}$ 
                        & $\textbf{63.18}$ \\  
    \Xhline{3\arrayrulewidth}
    \end{tabular}
}
\end{center}
\caption{ Test Accuracy for symmetric noise (Equation~\ref{eqn:sym_noise}) removal from CIFAR10 and CIFAR100 datasets trained on VGG11 and  symmetric noise, asymmetric noise (Equation~\ref{eqn:asym_noise}) and hierarchical noise (Equation~\ref{eqn:real_noise}) removal ResNet18 architectures with noise strength $\eta=0.1$ and $\eta=0.25$. Note, the confusing class groups (or set $\mathcal{C}$) for hierarchical noise are cat-dog and truck-automobile for CIFAR10. For CIFAR100 we use Superclasses as the confusing groups.
\textbf{\textsuperscript{\textdagger}} - We select 1000 samples with low loss value from $\mathcal{D}_{Tr}$ as retain samples for our method. 
} \label{tab:cifar}
\end{table*}

\subsection{Types of Synthetic Noise} 
We evaluate our approach under various noise settings, characterized by the noise transition matrix, denoted by $\mathcal{T}$. 
The noise transition matrix $\mathcal{T}$ is a square matrix of size $K \times K$, where $K$ is the number of classes, which captures the conditional probability distribution of label corruption. 
The element at the $i^{th}$ row and $j^{th}$ column, denoted by $t_{ij}$, describes the probability of a clean label, $c_i$, being flipped to a noisy label $c_j$. 
The degree of noise added is controlled by the parameter $\eta$. 
We study three different noise setting as described below:
    \subsubsection{Symmetric Noise: } 
    In symmetric noise \cite{ghosh2017robustlossfunctionslabel}, labels are flipped to any other class with equal probability. This creates a uniform distribution of errors across all classes. Equation~\ref{eqn:sym_noise} defines the element of the noise transition matrix $\mathcal{T}$ under a  symmetric noise setting. 
    \begin{align}\label{eqn:sym_noise}
        \begin{split}
            t_{ij}  = 
            \begin{cases}
                               \frac{\eta}{(K-1)} \qquad &i \neq j\\
                               1-\eta  & i = j            
            \end{cases}
        \end{split}
    \end{align}
    \subsubsection{Asymmetric Noise: } In asymmetric noise \cite{ghosh2017robustlossfunctionslabel}, the probability of a label being flipped to a specific class varies. The noise transition matrix $\mathcal{T}$ in our experiments is defined by Equation~\ref{eqn:asym_noise}.
    The values for off-diagonal elements are sampled from a uniform distribution $\mathcal{U}$ over the range 0 to $\frac{2\eta}{(K-1)}$. 
    The diagonal elements, $t_{ii}$, are computed to ensure that each row of the matrix sums to 1.    
    \begin{align}\label{eqn:asym_noise}
    \vspace{-3mm}
        \begin{split}
            t_{ij}  = 
            \begin{cases}
                               \epsilon \sim \mathcal{U}(0,\frac{2\eta}{(K-1)}) \qquad &i \neq j\\
                               \\
                               1 - \sum_{k=1; k\neq i}^{k=K} t_{ik}  & i = j            
            \end{cases}
        \end{split}
    \end{align}
    \subsubsection{Hierarchical Noise: } To simulate a practical label noise scenario, we introduce label errors based on class hierarchy \cite{class_hierarchy}. 
    This approach reflects the intuition that human labelers are more likely to confuse hierarchically closer classes. 
    We define a set of confusing class groups, $\mathcal{C}$, where $g_i \in \mathcal{C}$, indicates all the classes which are hierarchically closer to class $i$.
     Equation~\ref{eqn:real_noise} represent the noise transition matrix for this case.     
    \begin{align}\label{eqn:real_noise}
        \begin{split}
            t_{ij}  = 
            \begin{cases}
                               \frac{\eta}{|g_i|} \quad &i\neq j; \hspace{2mm} j\in g_i \\
                               0  &i\neq j;  \hspace{2mm} j\notin g_i \\
                               1 - \sum_{k=1; k\neq i}^{k=K} t_{ik}  & i = j        
            \end{cases}\\
        \text{ where }g_i \in \mathcal{C}; \text{ and } |g_i| \text{ number of classes in }g_i
        \end{split}
    \end{align}

\subsection{Corrective Machine Unlearning Benchmark}
We compare our proposed method, {\bf\em SAP}, to several unlearning algorithms under different synthetic label noise listed above. 
As a reference point, we include a {\bf Retrain} model trained with Stochastic Gradient Descent (SGD) on the clean data partition, $\mathcal{D}^{cln}_{Tr}$.
All unlearning methods are applied to the {\bf Vanilla} model, which is trained with SGD on the entire corrupt dataset, $\mathcal{D}_{Tr}$.
We also include a {\bf Finetune} baseline, which updates all layers of the Vanilla model using SGD on a small retained dataset, leveraging the idea of catastrophic forgetting as explored in CF-k \cite{cfk}.
For a strong baseline, we implement {\bf SCRUB} \cite{scrub}, a state-of-the-art unlearning algorithm, and {\bf SSD} \cite{ssd}, which, like SAP, uses a single update for unlearning.

Many unlearning algorithms require access to both retain (correctly labeled) and forget (misclassified) samples \cite{goel2024correctivemachineunlearning}.
However, accurately distinguishing between hard and misclassified samples is challenging \cite{garg2023memorization}. 
To address this, we provide all unlearning algorithms with small subsets of retain and forget samples for evaluation. 
Unlike other methods, SAP does not explicitly use forget samples for model correction and uses the sample loss to identify potential samples for the retain set, bypassing the need for curated subsets. 
We observed other unlearning algorithms suffered significant performance degradation when using samples with low loss as the retained set.

\subsubsection{Results:} 
Our results in Table~\ref{tab:cifar} show that Retrain models closely match expected generalization performance, indicating that a slight reduction in dataset size has minimal impact on model generalization \cite{guo2022deepcore}. 
In contrast, the Vanilla model experiences a significant drop in test accuracy, especially at higher noise levels, with a $15-20\%$ decline observed at $25\%$ corruption level for both CIFAR10 and CIFAR100. 
This highlights the detrimental impact of label noise on performance.
Finetuning the Vanilla model for 50 epochs on a small subset of $\mathcal{D}_{Tr}^{cln}$ (the clean partition of $\mathcal{D}_{Tr}$) containing 5000 randomly selected samples consistently improves performance, with a notable $6\%$ gain for a ResNet18 model on CIFAR10 with $25\%$ symmetric label corruption.
Interestingly, SSD-based unlearning fails to provide a consistent generalization improvement, which might be attributed to the small sample size, as observed in previous studies \cite{goel2024correctivemachineunlearning,kodge2024deep}. 
{\em SAP} outperforms all other unlearning methods on both CIFAR-10 and CIFAR-100 datasets, achieving an average improvement of $1.36\%$ and $0.39\%$ compared to the second-best method, respectively. 
In our experiments, we observed that increasing the number of retain and forget samples to 5000 and 1000, respectively, improved the performance of the SCRUB algorithm, as presented in Table~4 in the supplementary material.
However, SCRUB requires extensive hyperparameter tuning (retain batch size, forget batch size, $\alpha$, $\gamma$, number of minimization steps, and number of maximization steps) to counteract the instability of gradient ascent, leading to approximately 675 hyperparameter search iterations compared to our method's 16. 
Increasing the number of samples further exacerbates the computational burden for hyperparameter search. 
In the next subsection, we demonstrate how SAP can enhance the generalization performance of label noise robust learning algorithms.


\subsection{Noise-Robust Learning Benchmarks }
\begin{table}[t]
\begin{center}
\resizebox{0.9\columnwidth}{!}{
\begin{tabular}{l|l||cc||c}
\Xhline{3\arrayrulewidth}
& \multirow{1}{*}{\bf Method}  & {\bf Baseline}& {\bf SAP}& {\bf Improvements}\\
    \hline  
    \hline
    \multirow{5}{*}{\bf \rotatebox[origin=c]{90}{CIFAR10}}
        & Vanilla   &  $79.47\pm0.46$ &$85.46\pm0.41$&$\textbf{5.99}$\\ 
        & Logit Clip   &$82.91\pm0.32$ & $85.99\pm0.67$&$\textbf{3.08}$\\
        & MixUp  &$83.12\pm0.44$ & $86.45\pm0.52$&$\textbf{3.33}$\\
        & SAM &$83.29\pm0.28$ & $87.29\pm0.08$&$\textbf{4.0}$\\
        & MentorMix    &$89.64\pm0.32$ & $90.51\pm0.17$&$\textbf{0.87}$\\
        \Xhline{3\arrayrulewidth}
            \multicolumn{2}{c||}{\bf Average}&$83.69$ & $87.14$&$\textbf{3.45}$\\
        \Xhline{3\arrayrulewidth}
    \multirow{5}{*}{\bf \rotatebox[origin=c]{90}{CIFAR100}}
        & Vanilla   & $54.75\pm0.45$&$58.69\pm0.68$&$\textbf{3.94}$\\ 
        & Logit Clip   &$56.41\pm0.43$ & $59.90\pm0.84$&$\textbf{3.49}$\\
        & SAM &$56.49\pm0.93$ & $59.34\pm0.76$&$\textbf{2.85}$\\
        & MixUp  &$58.25\pm0.65$ & $62.32\pm0.91$&$\textbf{4.07}$\\
        & MentorMix    &$68.53\pm0.35$ & $68.98\pm0.45$&$\textbf{0.45}$\\
        \Xhline{3\arrayrulewidth}
            \multicolumn{2}{c||}{\bf Average}&$58.89$&$61.85$&$\textbf{2.98}$\\
        \Xhline{3\arrayrulewidth}
\end{tabular}
}
\end{center}
\caption{ Generalization benefits of applying our algorithm in synergy with noise-robust training approaches for symmetric noise (Equation~\ref{eqn:sym_noise}) removal from CIFAR10 and CIFAR100 datasets trained on ResNet18 architectures with noise strength $\eta=0.25$. We report the mean and standard deviation across three randomly chosen seeds.
} \label{tab:cifar_lnl}
\end{table}
We compare our method to several label noise robust learning algorithms:
\textbf{ Logit Clip} \cite{wei2023mitigatingmemorizationnoisylabels}, Sharpness Aware Minimization (\textbf{SAM}) \cite{foret2021sharpnessaware}, \textbf{MixUp} \cite{mixup}, and \textbf{MentorMix} \cite{mentormix}, known for their inherent tolerance to mislabeled data. 
Table~\ref{tab:cifar_lnl} presents the performance of various algorithms on CIFAR-10 and CIFAR-100 datasets using the ResNet18 model under symmetric label noise with $\eta=0.25$. Notably, SAP improves generalization performance by $3.45\%$ and $2.98\%$ on average for CIFAR-10 and CIFAR-100, respectively.

\subsection{Real World Noisy Benchmark}
To demonstrate {\em SAP}'s effectiveness in real-world noise scenarios, we present experiments on Mini-WebVision \cite{mentormix} and Clothing1M \cite{xiao2015learning} datasets. 
Mini-WebVision is trained on the InceptionResNetV2 (IRV2 in Table~\ref{tab:real_world_noise}) \cite{szegedy2017inception} architecture, while Clothing1M is trained on ResNet50 \cite{he2016deep} and ViT\_B\_16. 
Comprehensive training and hyperparameter details are provided in supplementary material.

As shown in Table~\ref{tab:real_world_noise}, applying {\em SAP} consistently yields generalization improvements for a noisy dataset. 
These results demonstrate an average improvement of $1.8\%$, highlighting {\em SAP}'s ability to scale to large datasets (Clothing1M) and SoTA transformer based models (ViT\_B\_16).  
\begin{table}[t]
    \begin{center}
    \resizebox{1.0\columnwidth}{!}{
        \begin{tabular}{l|l|cc||c}
        \Xhline{3\arrayrulewidth}
                            \multirow{1}{*}{\bf Dataset } & \multirow{1}{*}{\bf Architecture }& {\bf Vanilla} &{\bf SAP } &{\bf Improvement} \\
                        \hline  \hline 
                            Mini-WebVision &IRV2&$63.81\pm0.38$&$64.73\pm0.53$ & $\textbf{0.92}$\\
                            Clothing1M &ResNet50&$67.48\pm0.64$&$69.64\pm0.57$ & $\textbf{2.16}$\\
                            Clothing1M &ViT\_B\_16 &$69.12\pm0.45$&$71.43\pm0.60$& $\textbf{2.31}$\\
                        \Xhline{3\arrayrulewidth}
                            \multicolumn{2}{c|}{\bf Average}&$66.80$&$68.60$ & $\textbf{1.80}$\\
                        \Xhline{3\arrayrulewidth}
        \end{tabular}
                }
    \end{center}
    \caption{ Generalization benefits of applying our algorithm on real-world noisy dataset like Mini-WebVision \cite{mentormix}, and Clothing1M \cite{xiao2015learning} for Vanilla trained InceptionResNetv2, called IRV2 in Table \cite{szegedy2017inception} and ViT\_B\_16 \cite{vit}.} 
    \label{tab:real_world_noise}
\end{table}

\subsection{Analyses}
\subsubsection{Increasing Trusted Sample Size yields diminishing returns:}
We study how the number of trusted samples impacts {\em SAP}'s performance.  
Figure~\ref{fig:sample_size} shows how test accuracy varies with the number of trusted samples. 
We observe that increasing the sample size improves accuracy up to 1000 samples, after which there are no significant gains.  
Since representation calculations and SVD are performed on the trusted set, the sample size directly affects compute requirements.
Therefore, we use 1000 samples in our experiments to balance performance and computational efficiency.
\subsubsection{Optimal $\alpha$ is key to maximizing performance gains :}
\begin{figure}[htbp]
\centering
    \begin{minipage}{.48\columnwidth}
        \centering
        \includegraphics[width=1.0\linewidth, trim={1mm 3mm 2mm 0mm}, clip]{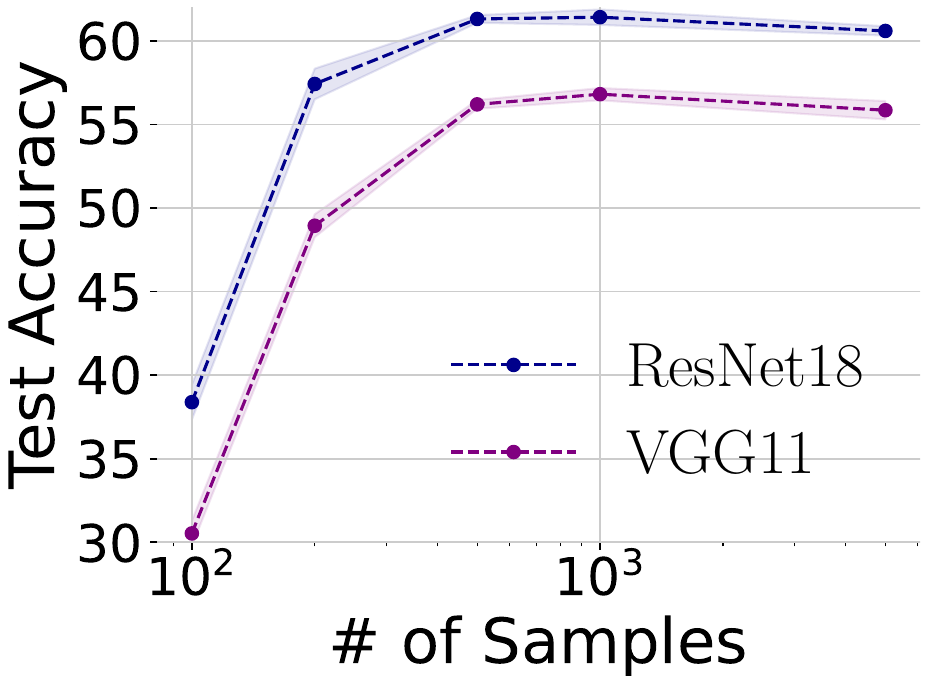}
        \caption{Effect of trusted sample size for CIFAR100 dataset with $\eta=25\%$.}
        \label{fig:sample_size}
    \end{minipage}
    \hspace{1mm}
    \begin{minipage}{.48\columnwidth}
        \centering
        \includegraphics[width=1.0\linewidth, trim={1mm 3mm 2mm 0mm}, clip]{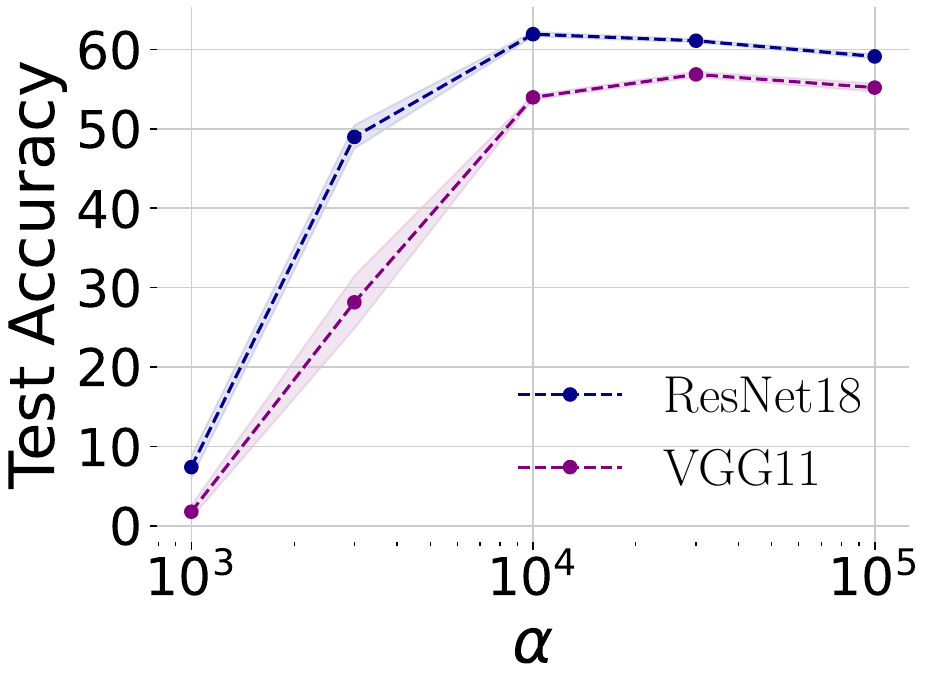}
        \caption{Effect of $\alpha$ for model trained on CIFAR100 dataset with $\eta=25\%$.}
        \label{fig:alpha_effect}
    \end{minipage} 
    \vspace{-4mm}
\end{figure}

Figure~\ref{fig:alpha_effect} demonstrates how the scaling factor ($\alpha$) influences test accuracy for models trained on datasets with $25\%$ label corruption. We observe that increasing $\alpha$ generally improves test accuracy up to a certain point, after which accuracy gradually declines. 
Importantly, we find that $\alpha=30000$ consistently yields strong results for synthetic noise across different datasets, models, and corruption levels. 
\section{Discussion and Conclusion}
\label{sec:conclusion}
This paper introduces {\em SAP}, a corrective machine unlearning algorithm that is competitive with the SoTA SCRUB without requiring explicit access to clean and mislabeled data subsets.
This eliminates the need for laborious human curation of training data.
{\em SAP} performs a single model update using Singular Value Decomposition (SVD) on a small sample set,
 ensuring computational efficiency.
Furthermore, with only one hyperparameter, $\alpha$, and reusable SVD computations across different $\alpha$ values, {\em SAP} offers simplicity and efficiency, unlike gradient-ascent based algorithms which often require extensive hyperparameter tuning,
 sometimes exceeding the computational cost of retraining the model. In summary, SAP is a
 sample and compute-efficient unlearning technique to address the challenges posed by noisy data.
\section{Acknowledgements}
This work was supported in part by the Center for the Co-Design of Cognitive Systems (COCOSYS), a DARPA-sponsored JUMP center, the Semiconductor Research Corporation (SRC), the National Science Foundation, and Collins Aerospace. 

\bibliography{main}

\begin{thebibliography}{55}
\providecommand{\natexlab}[1]{#1}

\bibitem[{Biggio, Nelson, and Laskov(2012)}]{biggo2012poisoning}
Biggio, B.; Nelson, B.; and Laskov, P. 2012.
\newblock Poisoning Attacks against Support Vector Machines.
\newblock In \emph{Proceedings of the 29th International Coference on International Conference on Machine Learning}, ICML'12, 1467–1474. Madison, WI, USA: Omnipress.
\newblock ISBN 9781450312851.

\bibitem[{Chen et~al.(2017)Chen, Liu, Li, Lu, and Song}]{chen2017targeted}
Chen, X.; Liu, C.; Li, B.; Lu, K.; and Song, D. 2017.
\newblock Targeted backdoor attacks on deep learning systems using data poisoning.
\newblock \emph{arXiv preprint arXiv:1712.05526}.

\bibitem[{Chen et~al.(2022)Chen, Shen, Shen, Wang, and Zhang}]{chen2022amplifying}
Chen, Y.; Shen, C.; Shen, Y.; Wang, C.; and Zhang, Y. 2022.
\newblock Amplifying membership exposure via data poisoning.
\newblock \emph{Advances in Neural Information Processing Systems}, 35: 29830--29844.

\bibitem[{Cheng et~al.(2022)Cheng, Liu, Ning, Wang, Han, Niu, Gao, and Sugiyama}]{Cheng_2022_CVPR}
Cheng, D.; Liu, T.; Ning, Y.; Wang, N.; Han, B.; Niu, G.; Gao, X.; and Sugiyama, M. 2022.
\newblock Instance-Dependent Label-Noise Learning With Manifold-Regularized Transition Matrix Estimation.
\newblock In \emph{Proceedings of the IEEE/CVF Conference on Computer Vision and Pattern Recognition (CVPR)}, 16630--16639.

\bibitem[{Cheng et~al.(2021)Cheng, Zhu, Li, Gong, Sun, and Liu}]{cheng2021learninginstancedependentlabelnoise}
Cheng, H.; Zhu, Z.; Li, X.; Gong, Y.; Sun, X.; and Liu, Y. 2021.
\newblock Learning with Instance-Dependent Label Noise: A Sample Sieve Approach.
\newblock arXiv:2010.02347.

\bibitem[{Chetlur et~al.(2014)Chetlur, Woolley, Vandermersch, Cohen, Tran, Catanzaro, and Shelhamer}]{chetlur2014cudnnefficientprimitivesdeep}
Chetlur, S.; Woolley, C.; Vandermersch, P.; Cohen, J.; Tran, J.; Catanzaro, B.; and Shelhamer, E. 2014.
\newblock cuDNN: Efficient Primitives for Deep Learning.
\newblock arXiv:1410.0759.

\bibitem[{Deisenroth, Faisal, and Ong(2020)}]{Deisenroth2020}
Deisenroth, M.~P.; Faisal, A.~A.; and Ong, C.~S. 2020.
\newblock \emph{Mathematics for Machine Learning}.
\newblock Cambridge University Press.

\bibitem[{Dosovitskiy et~al.(2021)Dosovitskiy, Beyer, Kolesnikov, Weissenborn, Zhai, Unterthiner, Dehghani, Minderer, Heigold, Gelly, Uszkoreit, and Houlsby}]{vit}
Dosovitskiy, A.; Beyer, L.; Kolesnikov, A.; Weissenborn, D.; Zhai, X.; Unterthiner, T.; Dehghani, M.; Minderer, M.; Heigold, G.; Gelly, S.; Uszkoreit, J.; and Houlsby, N. 2021.
\newblock An Image is Worth 16x16 Words: Transformers for Image Recognition at Scale.
\newblock arXiv:2010.11929.

\bibitem[{Feng, Quan, and Dauphin(2020)}]{feng2020label}
Feng, W.; Quan, Y.; and Dauphin, G. 2020.
\newblock Label noise cleaning with an adaptive ensemble method based on noise detection metric.
\newblock \emph{Sensors}, 20(23): 6718.

\bibitem[{Foret et~al.(2021)Foret, Kleiner, Mobahi, and Neyshabur}]{foret2021sharpnessaware}
Foret, P.; Kleiner, A.; Mobahi, H.; and Neyshabur, B. 2021.
\newblock Sharpness-aware Minimization for Efficiently Improving Generalization.
\newblock In \emph{International Conference on Learning Representations}.

\bibitem[{Foster, Schoepf, and Brintrup(2024)}]{ssd}
Foster, J.; Schoepf, S.; and Brintrup, A. 2024.
\newblock Fast machine unlearning without retraining through selective synaptic dampening.
\newblock In \emph{Proceedings of the AAAI Conference on Artificial Intelligence}, volume 38-11, 12043--12051.

\bibitem[{Garg, Ravikumar, and Roy(2024)}]{garg2023memorization}
Garg, I.; Ravikumar, D.; and Roy, K. 2024.
\newblock Memorization Through the Lens of Curvature of Loss Function Around Samples.
\newblock In \emph{Forty-first International Conference on Machine Learning}.

\bibitem[{Ghosh, Kumar, and Sastry(2017)}]{ghosh2017robustlossfunctionslabel}
Ghosh, A.; Kumar, H.; and Sastry, P.~S. 2017.
\newblock Robust Loss Functions under Label Noise for Deep Neural Networks.
\newblock arXiv:1712.09482.

\bibitem[{Goel et~al.(2022)Goel, Prabhu, Sanyal, Lim, Torr, and Kumaraguru}]{cfk}
Goel, S.; Prabhu, A.; Sanyal, A.; Lim, S.-N.; Torr, P.; and Kumaraguru, P. 2022.
\newblock Towards adversarial evaluations for inexact machine unlearning.
\newblock \emph{arXiv preprint arXiv:2201.06640}.

\bibitem[{Goel et~al.(2024)Goel, Prabhu, Torr, Kumaraguru, and Sanyal}]{goel2024correctivemachineunlearning}
Goel, S.; Prabhu, A.; Torr, P.; Kumaraguru, P.; and Sanyal, A. 2024.
\newblock Corrective Machine Unlearning.
\newblock arXiv:2402.14015.

\bibitem[{Guo, Zhao, and Bai(2022)}]{guo2022deepcore}
Guo, C.; Zhao, B.; and Bai, Y. 2022.
\newblock Deepcore: A comprehensive library for coreset selection in deep learning.
\newblock In \emph{International Conference on Database and Expert Systems Applications}, 181--195. Springer.

\bibitem[{Hayes et~al.(2024)Hayes, Shumailov, Triantafillou, Khalifa, and Papernot}]{hayes2024inexactunlearningneedscareful}
Hayes, J.; Shumailov, I.; Triantafillou, E.; Khalifa, A.; and Papernot, N. 2024.
\newblock Inexact Unlearning Needs More Careful Evaluations to Avoid a False Sense of Privacy.
\newblock arXiv:2403.01218.

\bibitem[{He et~al.(2016)He, Zhang, Ren, and Sun}]{he2016deep}
He, K.; Zhang, X.; Ren, S.; and Sun, J. 2016.
\newblock Deep residual learning for image recognition.
\newblock In \emph{Proceedings of the IEEE conference on computer vision and pattern recognition}, 770--778.

\bibitem[{Jagielski et~al.(2018)Jagielski, Oprea, Biggio, Liu, Nita-Rotaru, and Li}]{jagielski2018manipulating}
Jagielski, M.; Oprea, A.; Biggio, B.; Liu, C.; Nita-Rotaru, C.; and Li, B. 2018.
\newblock Manipulating machine learning: Poisoning attacks and countermeasures for regression learning.
\newblock In \emph{2018 IEEE symposium on security and privacy (SP)}, 19--35. IEEE.

\bibitem[{Jia et~al.(2022)Jia, Li, Yu, Bian, Zhao, Li, Xiong, and Dou}]{jia2022learningtrainingdynamicsidentifying}
Jia, Q.; Li, X.; Yu, L.; Bian, J.; Zhao, P.; Li, S.; Xiong, H.; and Dou, D. 2022.
\newblock Learning from Training Dynamics: Identifying Mislabeled Data Beyond Manually Designed Features.
\newblock arXiv:2212.09321.

\bibitem[{Jiang et~al.(2020)Jiang, Huang, Liu, and Yang}]{mentormix}
Jiang, L.; Huang, D.; Liu, M.; and Yang, W. 2020.
\newblock Beyond synthetic noise: Deep learning on controlled noisy labels.
\newblock In \emph{International conference on machine learning}, 4804--4815. PMLR.

\bibitem[{Jiang et~al.(2018)Jiang, Zhou, Leung, Li, and Fei-Fei}]{jiang2018mentornet}
Jiang, L.; Zhou, Z.; Leung, T.; Li, L.-J.; and Fei-Fei, L. 2018.
\newblock Mentornet: Learning data-driven curriculum for very deep neural networks on corrupted labels.
\newblock In \emph{International conference on machine learning}, 2304--2313. PMLR.

\bibitem[{Jindal, Nokleby, and Chen(2016)}]{jindal2016learning}
Jindal, I.; Nokleby, M.; and Chen, X. 2016.
\newblock Learning deep networks from noisy labels with dropout regularization.
\newblock In \emph{2016 IEEE 16th International Conference on Data Mining (ICDM)}, 967--972. IEEE.

\bibitem[{Kodge, Saha, and Roy(2024)}]{kodge2024deep}
Kodge, S.; Saha, G.; and Roy, K. 2024.
\newblock Deep Unlearning: Fast and Efficient Gradient-free Class Forgetting.
\newblock \emph{Transactions on Machine Learning Research}.

\bibitem[{Krizhevsky, Hinton et~al.(2009)}]{cifar}
Krizhevsky, A.; Hinton, G.; et~al. 2009.
\newblock Learning multiple layers of features from tiny images.
\newblock \emph{Technical Report}.

\bibitem[{Kurmanji, Triantafillou, and Triantafillou(2023)}]{scrub}
Kurmanji, M.; Triantafillou, P.; and Triantafillou, E. 2023.
\newblock Towards Unbounded Machine Unlearning.
\newblock \emph{arXiv preprint arXiv:2302.09880}.

\bibitem[{Li, Socher, and Hoi(2020)}]{Li2020DivideMix}
Li, J.; Socher, R.; and Hoi, S.~C. 2020.
\newblock DivideMix: Learning with Noisy Labels as Semi-supervised Learning.
\newblock In \emph{International Conference on Learning Representations}.

\bibitem[{Lukasik et~al.(2020)Lukasik, Bhojanapalli, Menon, and Kumar}]{lukasik2020does}
Lukasik, M.; Bhojanapalli, S.; Menon, A.; and Kumar, S. 2020.
\newblock Does label smoothing mitigate label noise?
\newblock In \emph{International Conference on Machine Learning}, 6448--6458. PMLR.

\bibitem[{Maini et~al.(2022)Maini, Garg, Lipton, and Kolter}]{maini2022characterizing}
Maini, P.; Garg, S.; Lipton, Z.; and Kolter, J.~Z. 2022.
\newblock Characterizing datapoints via second-split forgetting.
\newblock \emph{Advances in Neural Information Processing Systems}, 35: 30044--30057.

\bibitem[{Mukherjee, Garg, and Roy(2024)}]{class_hierarchy}
Mukherjee, A.; Garg, I.; and Roy, K. 2024.
\newblock Encoding Hierarchical Information in Neural Networks Helps in Subpopulation Shift.
\newblock \emph{IEEE Transactions on Artificial Intelligence}, 5(2): 827--838.

\bibitem[{Northcutt, Jiang, and Chuang(2021)}]{northcutt2021confident}
Northcutt, C.; Jiang, L.; and Chuang, I. 2021.
\newblock Confident learning: Estimating uncertainty in dataset labels.
\newblock \emph{Journal of Artificial Intelligence Research}, 70: 1373--1411.

\bibitem[{Northcutt, Athalye, and Mueller(2021)}]{northcutt2021pervasive}
Northcutt, C.~G.; Athalye, A.; and Mueller, J. 2021.
\newblock Pervasive Label Errors in Test Sets Destabilize Machine Learning Benchmarks.
\newblock In \emph{Thirty-fifth Conference on Neural Information Processing Systems Datasets and Benchmarks Track (Round 1)}.

\bibitem[{Pangakis and Wolken(2024)}]{pangakis2024knowledgedistillationautomatedannotation}
Pangakis, N.; and Wolken, S. 2024.
\newblock Knowledge Distillation in Automated Annotation: Supervised Text Classification with LLM-Generated Training Labels.
\newblock arXiv:2406.17633.

\bibitem[{Patel and Sastry(2023)}]{Patel_2023_WACV}
Patel, D.; and Sastry, P.~S. 2023.
\newblock Adaptive Sample Selection for Robust Learning Under Label Noise.
\newblock In \emph{Proceedings of the IEEE/CVF Winter Conference on Applications of Computer Vision (WACV)}, 3932--3942.

\bibitem[{Ravikumar et~al.(2024)Ravikumar, Soufleri, Hashemi, and Roy}]{ravikumar2024unveiling}
Ravikumar, D.; Soufleri, E.; Hashemi, A.; and Roy, K. 2024.
\newblock Unveiling Privacy, Memorization, and Input Curvature Links.
\newblock In \emph{Forty-first International Conference on Machine Learning}.

\bibitem[{Saha, Garg, and Roy(2021)}]{saha2021gradient}
Saha, G.; Garg, I.; and Roy, K. 2021.
\newblock Gradient Projection Memory for Continual Learning.
\newblock In \emph{International Conference on Learning Representations}.

\bibitem[{Sambasivan et~al.(2021)Sambasivan, Kapania, Highfill, Akrong, Paritosh, and Aroyo}]{sambasivan2021everyone}
Sambasivan, N.; Kapania, S.; Highfill, H.; Akrong, D.; Paritosh, P.; and Aroyo, L.~M. 2021.
\newblock “Everyone wants to do the model work, not the data work”: Data Cascades in High-Stakes AI.
\newblock In \emph{proceedings of the 2021 CHI Conference on Human Factors in Computing Systems}, 1--15.

\bibitem[{Schoepf, Foster, and Brintrup(2024)}]{assd}
Schoepf, S.; Foster, J.; and Brintrup, A. 2024.
\newblock Parameter-tuning-free data entry error unlearning with adaptive selective synaptic dampening.
\newblock arXiv:2402.10098.

\bibitem[{Schwarzschild et~al.(2021)Schwarzschild, Goldblum, Gupta, Dickerson, and Goldstein}]{schwarzschild2021just}
Schwarzschild, A.; Goldblum, M.; Gupta, A.; Dickerson, J.~P.; and Goldstein, T. 2021.
\newblock Just how toxic is data poisoning? a unified benchmark for backdoor and data poisoning attacks.
\newblock In \emph{International Conference on Machine Learning}, 9389--9398. PMLR.

\bibitem[{Simonyan and Zisserman(2015)}]{vgg}
Simonyan, K.; and Zisserman, A. 2015.
\newblock Very Deep Convolutional Networks for Large-Scale Image Recognition.
\newblock arXiv:1409.1556.

\bibitem[{Steinhardt, Koh, and Liang(2017)}]{steinhardt2017certified}
Steinhardt, J.; Koh, P. W.~W.; and Liang, P.~S. 2017.
\newblock Certified defenses for data poisoning attacks.
\newblock \emph{Advances in neural information processing systems}, 30.

\bibitem[{Szegedy et~al.(2017)Szegedy, Ioffe, Vanhoucke, and Alemi}]{szegedy2017inception}
Szegedy, C.; Ioffe, S.; Vanhoucke, V.; and Alemi, A. 2017.
\newblock Inception-v4, inception-resnet and the impact of residual connections on learning.
\newblock In \emph{Proceedings of the AAAI conference on artificial intelligence}, volume 31-1.

\bibitem[{Triantafillou et~al.(2024)Triantafillou, Kairouz, Pedregosa, Hayes, Kurmanji, Zhao, Dumoulin, Junior, Mitliagkas, Wan, Hosoya, Escalera, Dziugaite, Triantafillou, and Guyon}]{triantafillou2024makingprogressunlearningfindings}
Triantafillou, E.; Kairouz, P.; Pedregosa, F.; Hayes, J.; Kurmanji, M.; Zhao, K.; Dumoulin, V.; Junior, J.~J.; Mitliagkas, I.; Wan, J.; Hosoya, L.~S.; Escalera, S.; Dziugaite, G.~K.; Triantafillou, P.; and Guyon, I. 2024.
\newblock Are we making progress in unlearning? Findings from the first NeurIPS unlearning competition.
\newblock arXiv:2406.09073.

\bibitem[{Verma et~al.(2019)Verma, Lamb, Beckham, Najafi, Mitliagkas, Lopez-Paz, and Bengio}]{manifoldmixup}
Verma, V.; Lamb, A.; Beckham, C.; Najafi, A.; Mitliagkas, I.; Lopez-Paz, D.; and Bengio, Y. 2019.
\newblock Manifold mixup: Better representations by interpolating hidden states.
\newblock In \emph{International conference on machine learning}, 6438--6447. PMLR.

\bibitem[{Wang et~al.(2024)Wang, Kim, Rahman, Mitra, and Miao}]{wang2024human}
Wang, X.; Kim, H.; Rahman, S.; Mitra, K.; and Miao, Z. 2024.
\newblock Human-LLM collaborative annotation through effective verification of LLM labels.
\newblock In \emph{Proceedings of the CHI Conference on Human Factors in Computing Systems}, 1--21.

\bibitem[{Wang et~al.(2019)Wang, Ma, Chen, Luo, Yi, and Bailey}]{wang2019symmetric}
Wang, Y.; Ma, X.; Chen, Z.; Luo, Y.; Yi, J.; and Bailey, J. 2019.
\newblock Symmetric Cross Entropy for Robust Learning with Noisy Labels.
\newblock arXiv:1908.06112.

\bibitem[{Wei et~al.(2023)Wei, Zhuang, Xie, Feng, Niu, An, and Li}]{wei2023mitigatingmemorizationnoisylabels}
Wei, H.; Zhuang, H.; Xie, R.; Feng, L.; Niu, G.; An, B.; and Li, Y. 2023.
\newblock Mitigating Memorization of Noisy Labels by Clipping the Model Prediction.
\newblock arXiv:2212.04055.

\bibitem[{Wei et~al.(2021)Wei, Liu, Liu, Niu, Sugiyama, and Liu}]{wei2021smooth}
Wei, J.; Liu, H.; Liu, T.; Niu, G.; Sugiyama, M.; and Liu, Y. 2021.
\newblock To smooth or not? when label smoothing meets noisy labels.
\newblock \emph{arXiv preprint arXiv:2106.04149}.

\bibitem[{Xia et~al.(2020)Xia, Liu, Han, Gong, Wang, Ge, and Chang}]{xia2020robust}
Xia, X.; Liu, T.; Han, B.; Gong, C.; Wang, N.; Ge, Z.; and Chang, Y. 2020.
\newblock Robust early-learning: Hindering the memorization of noisy labels.
\newblock In \emph{International conference on learning representations}.

\bibitem[{Xiao et~al.(2015)Xiao, Xia, Yang, Huang, and Wang}]{xiao2015learning}
Xiao, T.; Xia, T.; Yang, Y.; Huang, C.; and Wang, X. 2015.
\newblock Learning from massive noisy labeled data for image classification.
\newblock In \emph{Proceedings of the IEEE conference on computer vision and pattern recognition}, 2691--2699.

\bibitem[{Zhang et~al.(2017)Zhang, Cisse, Dauphin, and Lopez-Paz}]{mixup}
Zhang, H.; Cisse, M.; Dauphin, Y.~N.; and Lopez-Paz, D. 2017.
\newblock mixup: Beyond empirical risk minimization.
\newblock \emph{arXiv preprint arXiv:1710.09412}.

\bibitem[{Zhang and Sabuncu(2018)}]{zhang2018generalized}
Zhang, Z.; and Sabuncu, M. 2018.
\newblock Generalized cross entropy loss for training deep neural networks with noisy labels.
\newblock \emph{Advances in neural information processing systems}, 31.

\bibitem[{Zheng, Awadallah, and Dumais(2021)}]{zheng2021meta}
Zheng, G.; Awadallah, A.~H.; and Dumais, S. 2021.
\newblock Meta label correction for noisy label learning.
\newblock In \emph{Proceedings of the AAAI conference on artificial intelligence}, volume 35-12, 11053--11061.

\bibitem[{Zheng et~al.(2020)Zheng, Wu, Goswami, Goswami, Metaxas, and Chen}]{pmlr-v119-zheng20c}
Zheng, S.; Wu, P.; Goswami, A.; Goswami, M.; Metaxas, D.; and Chen, C. 2020.
\newblock Error-Bounded Correction of Noisy Labels.
\newblock In III, H.~D.; and Singh, A., eds., \emph{Proceedings of the 37th International Conference on Machine Learning}, volume 119 of \emph{Proceedings of Machine Learning Research}, 11447--11457. PMLR.

\bibitem[{Zhu, Wang, and Liu(2022)}]{pmlr-v162-zhu22k}
Zhu, Z.; Wang, J.; and Liu, Y. 2022.
\newblock Beyond Images: Label Noise Transition Matrix Estimation for Tasks with Lower-Quality Features.
\newblock In Chaudhuri, K.; Jegelka, S.; Song, L.; Szepesvari, C.; Niu, G.; and Sabato, S., eds., \emph{Proceedings of the 39th International Conference on Machine Learning}, volume 162 of \emph{Proceedings of Machine Learning Research}, 27633--27653. PMLR.

\end{thebibliography}

\clearpage
\onecolumn
\appendix
\section{Experimental Setup for Toy Illustration} \label{apx:demo} 
In Figure~\ref{fig:boundary} we demonstrate our unlearning algorithm on a 2 way classification problem, where the original model is trained to detect samples from 2 different 2-dimensional spirals marked by green and blue points in Figure~\ref{fig:boundary_clean_dataset}. Below we present the details of the experiment. 

\subsubsection{Dataset: }
The training dataset has 250 samples per class and the test dataset has 5000 samples per class. 
\subsubsection{Model Architecture: }
We use a simple linear model with 10 layers with with ReLU activation functions. 
All the intermediate layers have 500 neurons and each layer excluding the final layer is followed by BatchNorm. 

\subsubsection{Training details: }
We train this network with stochastic gradient descent for 250 epochs with a learning rate of 0.01. We use learning rate scheduler which scales the learning rate by 0.7 when the training loss has stopped reducing. For these experiment we use a training batch size of 512.


\section{Brief Literature Survey on Label Noise Learning} \label{apx:survey} 
\begin{figure*}[htbp]
\label{fig:Label_Noise_Algorithm}
\centering
    \subfigure[ Dataset cleaning]
        {
        \label{fig:data_cleaning}
        \includegraphics[width=0.32\textwidth]{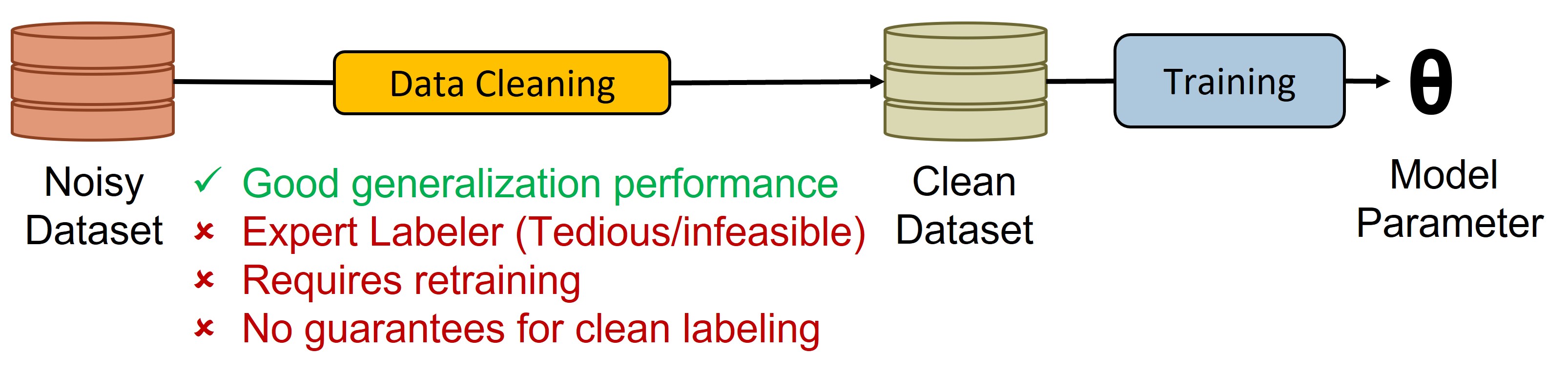}
        }
    \subfigure[ Noise-Robust Training]
        {
        \label{fig:robust_algorithm}
        \includegraphics[width=0.32\textwidth]{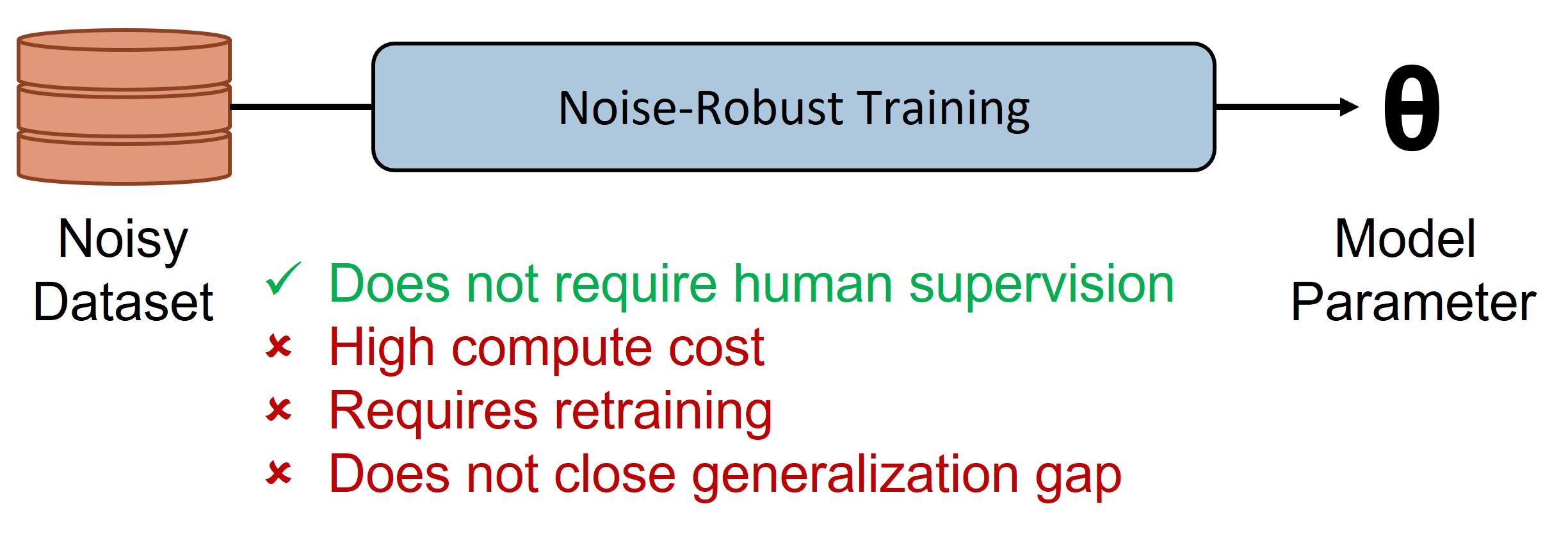}
        }
    \subfigure[Corrective Unlearning]
        {
        \label{fig:proposed}
        \includegraphics[width=0.32\textwidth]{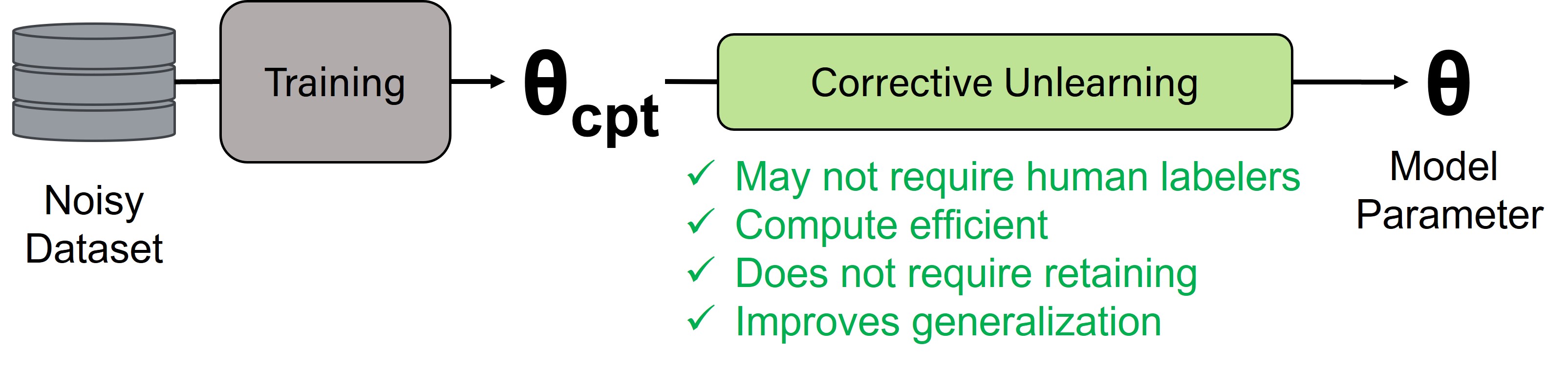}
        }
\caption{ Overview of Label Noise Handling. (a) Explores Data Cleaning approaches, (b) Presents Noise-Robust Training methods and (c) shows Corrective Unlearning. }
\end{figure*}
\textit{ Data Cleaning: } One prominent direction to handle label noise involves \textbf{data filtering}, where the goal is to identify and remove mislabeled samples within the training data.
Ensemble learning approaches \cite{feng2020label} have been explored for filtering. However, these methods can be computationally expensive due to the need to train multiple models.
To address this challenge, recent studies have focused on leveraging training dynamics to identify label noise, including methods using learning time \cite{xia2020robust}, forgetting time \cite{maini2022characterizing}, train-set confidence \cite{northcutt2021confident} or input loss curvature \cite{garg2023memorization,ravikumar2024unveiling}. 
These offline filtering often require the retraining of the network on the clean dataset. 
Researchers have creatively solved the above challenge with \textbf{sample selection} \cite{cheng2021learninginstancedependentlabelnoise,Patel_2023_WACV} techniques, where the algorithm dynamically removes samples from the training data. 
Since both filtering and sample selection can reduce training data size, potentially impacting model performance, research has also explored \textbf{label correction} \cite{pmlr-v119-zheng20c,zheng2021meta} algorithms.

\textit{ Noise-Robust Training: } Several works focus on developing algorithms that are inherently less sensitive to label noise. \textbf{Regularization} techniques, such as dropout \cite{jindal2016learning}, label smoothing \cite{lukasik2020does,wei2021smooth} or clipping the model predictions \cite{wei2023mitigatingmemorizationnoisylabels}, help prevent overfitting to noisy labels, leading to improved robustness.
\textbf{Noise robust loss functions} like symmetric cross-entropy \cite{wang2019symmetric} and Mean Absolute Error (MAE) less sensitive to outliers caused by noisy labels compared to standard cross-entropy loss.
Further, \textbf{Optimization} methods like Sharpness-Aware Minimization (SAM) \cite{foret2021sharpnessaware} promote flatter loss landscapes, which are associated with better generalization under label noise. 
Additionally, \textbf{curriculum learning} \cite{jiang2018mentornet,zhang2018generalized} gradually increase the difficulty of training examples, helping the model learn robust features before encountering noise. 
\textbf{Data augmentation} techniques such as MixUp \cite{mixup}, DivideMix \cite{Li2020DivideMix}, MentorMix\cite{mentormix} smooth training distributions, encouraging less abrupt decision boundaries and further enhancing robustness. 
Further, several studies estimate the {\bf Noise Transition Matrix} \cite{pmlr-v162-zhu22k,Cheng_2022_CVPR} to incorporate the label noise into the training process. 
While these robust algorithms offer improvement over standard training, however, do not completely close the generalization gap compared to the model trained on clean data. 

 \clearpage
\section{Representation Matrix for convolution layer} \label{apx:conv_representation} 
\begin{figure*}[htbp]
\centering     
\includegraphics[trim={3cm 9.0cm 5cm 6cm},clip,width=0.6\textwidth]{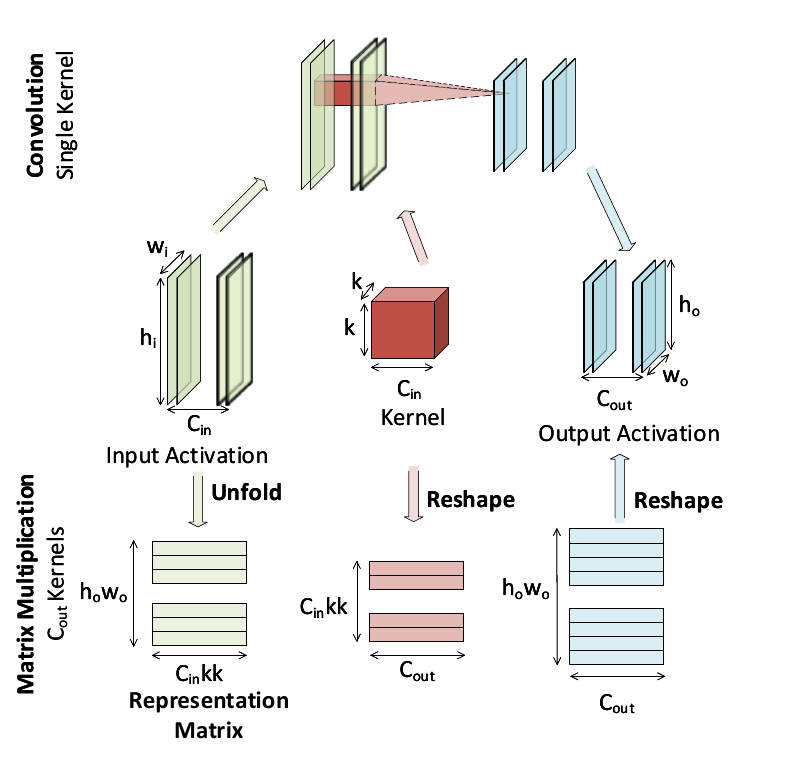}
\caption{ Representing convolution as matrix multiplication to enable weight update for unlearning. Each convolution kernel operates on patches of size $C_{in}\times k\times k$ in a sliding window fashion to produce an output map of size $h_o\times w_o$. The total number of patches is hence $n_p=h_ow_o$ for each pixel in the output activation map. The Unfold operation reshapes all these input patches for each sliding window to give a matrix of size $h_ow_o\times C_{in}kk$.  Unfolding the input activation and reshaping the weight kernel mathematically converts convolution into matrix multiplication operation as shown in the figure. The Unfolded input activations are collected as Representation Matrix for estimating Spaces using SVD. }
\label{fig:conv_rep}
\end{figure*}

Figure~\ref{fig:conv_rep} illustrates the unfold operations in details. 

 \clearpage
\section{Training and Hyperparameter Details}
\label{apx:training_details}

\subsection{Sythetic Noise with CIFAR Dataset}
For our experiments on CIFAR10 and CIFAR100, we use modified versions of ResNet18 and VGG11 with BatchNorm.  All the hyperparameters ( or hyperparameter tuning ranges ) are kept identical for both datasets. Below we present the hyperparameter details for several baselines. 

\subsubsection{Ideal/Retrain/Vanilla Baseline :} 
We train our models using Stochastic Gradient Descent (SGD) with an initial learning rate of $0.01$. We employ a weight decay of $5\times 10^{-4}$ with Nesterov accelerated momentum with the value of $0.9$. 
Our experiments on these datasets use a batch size of 64.

\subsubsection{Finetune Baseline:} 
We randomly sample 5000 data points from the clean partition of the dataset ($\mathcal{D}_{Tr}^{cln}$) to finetune the Vanilla model using the Post-Training Correction paradigm. For this fine-tuning baseline, we use SGD algorithm for 50 epochs and tune the learning rate using three values [0.005, 0.001, 0.0005].
We also tried using our loss-based sampling technique for finetuning as our method, however, we found choosing low loss samples catastrophically decreases the generalization performance of the model. 

\subsubsection{SSD:} SSD has the hyperparameter $\alpha$ which is tuned in the ranges [0.1, 1, 10, 50, 100, 500, 1000, 1e4, 1e5, 1e6] and hyperparameter $\lambda = \beta \times \alpha$ where $\beta$ is in the range [0.1, 0.5, 1.0, 5, 10] following \cite{goel2024correctivemachineunlearning}.

\subsubsection{SCRUB:} We tune the following hyperparameter for SCRUB:

\setlength{\itemindent}{10pt}
\begin{itemize}
    \item retain-batch-size - [32, 64, 128]
    \item forget-batch-size - [32, 64, 128]
    \item $\alpha$ - [0.03, 1, 0.3, 1, 3]
    \item $\gamma$ - [0.03]
    \item number of minimization steps - [5, 10,  15, 20, 25]
    \item number of maximization steps - [1, 3, 10]
\end{itemize}

\subsubsection{SAP:} We introduce the scaling hyperparameter $\alpha$ (Equation~\ref{eqn:scaling}) and perform a search within the range [2000, 4000, 8000, 10000, 12500, 15000, 17500, 20000, 22500, 25000, 30000, 40000, 50000, 75000, 100000, 300000].

\subsubsection{MixUp :} MixUp introduces a hyperparameter $\alpha$ controlling the mixing of images. We tune the hyperparameter in [0.1, 0.2, 0.4, and 0.8 ]. We find $\alpha=0.2$ works the best for both datasets. 

\subsubsection{MentorMix :} We tune the hyperparameter $\gamma_p$ in [0.3, 0.6, 0.7, 0.85] and find $\gamma_p=0.85$ works the best. We use $\alpha=0.2$ for MentorMix.

\subsubsection{SAM :} We tune the hyperparameter $\rho$ for SAM in [0.05, 0.1] and found $0.05$ gave better results.

\subsubsection{Logit Clip:} We tune the hyperparameter temperature for logit clipping in [0.01, 0.02, 0.04, 0.06, 0.08, 1.0, 1.5, 2, 2.5, 3.0, 3.5, 4.0, 4.5, 5.0] and found $3.0$ gave better results.

\subsection{Real World Noise}
For our experiments on Mini-WebVision we use InceptionResNetV2.  Experiments on Clothing1M are presented on ResNet50 and ViT\_B\_16 architecture.
Below we present the hyperparameter details for several baselines. 

\subsubsection{Vanilla :} For convolutional networks, we train our models using Stochastic Gradient Descent (SGD) with an initial learning rate of $0.01$. We employ a weight decay of $4\times 10^{-5}$ with Nesterov  accelerated momentum with the value of $0.9$. Our experiments on these datasets use a batch size of 64 for Mini-WebVision, and Clothing1M.
We randomly initialize the network and train the networks for 200 and 60 epochs for Mini-WebVision and WebVision1.0, respectively. For Clothing1M we initialized the network with Pytorch ResNet50 model pretrained on ImageNet1k. 
We keep these hyperparameters the same for the rest of the baselines. 

For the transformer based model (ViT\_B\_16), we use the learning rate of $2^{-5}$ and weight decay of 0.01 with a nestrove momentum of 0.9. We initialize the network with the pretrained ImageNet1k weights. We use the batchsize of 1024 and train the model for 10 epochs. 

\subsubsection{SAP :} We tune the scaling hyperparameter $\alpha$ (Equation~\ref{eqn:scaling})  within the range [$5e^{3}$, $1e^{4}$, $2e^{4}$, $3e^{4}$, $4e^{4}$, $6e^{4}$, $8e^{4}$, $1e^{5}$, $2e^{5}$, $4e^{5}$, $6e^{5}$, $8e^{5}$, $1e^{6}$, $2e^{6}$, $4e^{6}$, $6e^{6}$ ].

\clearpage
\section{Additional Results for SCRUB}
{\em SAP} outperforms all unlearning methods on both CIFAR10 and CIFAR100 datasets with and average improvement of $1.36\%$ and $0.39\%$ in comparison to the second best method as presented in Table~\ref{tab:cifar}. We observe that increasing the number of retain samples to 5000 and forget samples to 1000 increases the performance of SCRUB algorithm making SCRUB competitive with SAP for CIFAR10 and marginally ($0.71\%$) superior to SAP for CIFAR100 dataset. 
However, SCRUB requires extensive hyperparameter tuning (retain batch size, forget batch size, $\alpha$, $\gamma$, number of minimization steps, and number of maximization steps) to counteract the instability of gradient ascent, leading to approximately 675 hyperparameter search iterations compared to our method's 16. 
Increasing the number of samples further exacerbates the computational burden for hyperparameter search. 
\begin{table*}[htbp]
\begin{center}
\resizebox{1.0\textwidth}{!}{
    \begin{tabular}{l|l|c|c||cc||cc|cc|cc||c}
    \Xhline{3\arrayrulewidth}
    & \multirow{3}{*}{\bf Method} & & &  \multicolumn{2}{c||}{\bf \texttt{VGG11\_BN}} &  \multicolumn{6}{c||}{\bf \texttt{ResNet18}}&\multirow{3}{*}{\bf Average} \\
            \cline{5-12}
    &&{\bf Retain} &{\bf Forget}  &\multicolumn{2}{c||}{\bf Symmetric Noise ( Eqn.~\ref{eqn:sym_noise} )} &  \multicolumn{2}{c|}{\bf Symmetric Noise ( Eqn.~\ref{eqn:sym_noise} )} &  \multicolumn{2}{c|}{\bf Asymmetric Noise ( Eqn.~\ref{eqn:asym_noise} )} &  \multicolumn{2}{c||}{\bf Hierarchical Noise ( Eqn.~\ref{eqn:real_noise} )}\\
    && {\bf samples}& {\bf samples}& {\bf $\eta=0.1$}& {\bf $\eta=0.25$}& {\bf $\eta=0.1$}& {\bf $\eta=0.25$}& {\bf $\eta=0.1$}& {\bf $\eta=0.25$}& {\bf $\eta=0.1$}& {\bf $\eta=0.25$}\\
        \hline  
        \hline
        \multirow{3}{*}{\bf \rotatebox[origin=c]{30}{CIFAR10}}
            
            & SCRUB   & $1000$ & $200$  
                        & $85.88\pm0.35$&$78.90\pm0.25$
                        & $89.50\pm0.17$&$83.77\pm0.44$
                        & $89.50\pm0.22$&$83.60\pm0.14$
                        &$91.65\pm0.12$&$88.00\pm0.58$ 
                        & $86.35$\\
                        
            & SCRUB   & $5000$ & $1000$  
                        &$87.05\pm0.13$ & $81.60\pm0.39$ 
                        & $89.99\pm0.17$ & $85.33\pm0.30$ 
                        & $\textbf{90.19}\pm\textbf{0.18}$&$85.68\pm0.78$
                        &$\textbf{92.06}\pm\textbf{0.38}$&$\textbf{89.29}\pm\textbf{0.34}$ 
                        & $87.65$\\
            & SAP    & $0$\textsuperscript{\textdagger}  & $0$   
                        & $\textbf{87.25}\pm\textbf{0.16}$& $\textbf{82.27}\pm\textbf{0.15}$
                        & $\textbf{90.12}\pm\textbf{0.11}$&$\textbf{85.49}\pm\textbf{0.39}$
                        &$90.03\pm0.25$&$\textbf{86.32}\pm\textbf{0.66}$ 
                        &$91.87\pm0.22$&$87.92\pm0.69$ 
                        & $\textbf{87.66}$\\
    \hline
    \hline
        \multirow{3}{*}{\bf \rotatebox[origin=c]{30}{CIFAR100}}
            & SCRUB   & $1000$ & $200$  
                        & $60.93\pm0.09$&$52.11\pm0.63$
                        & $67.02\pm0.29$&$57.36\pm0.40$
                        & $73.12\pm0.18$&$63.37\pm0.69$
                        &$68.20\pm0.13$&$60.24\pm0.33$ 
                        &$62.79$ \\
            & SCRUB & $5000$ & $1000$       
                        &$\textbf{61.50}\pm\textbf{0.23}$ & $\textbf{54.54}\pm\textbf{0.96}$ 
                        & $\textbf{67.60}\pm\textbf{0.26}$ & $\textbf{59.23}\pm\textbf{0.44}$ 
                        & $\textbf{73.24}\pm\textbf{0.53}$&$\textbf{64.10}\pm\textbf{0.64}$
                        &$\textbf{68.83}\pm\textbf{0.28}$&$\textbf{62.10}\pm\textbf{0.06}$ 
                        &$\textbf{63.89}$ \\
            & SAP   & $0$\textsuperscript{\textdagger} & $0$ 
                        & $61.10\pm0.23$& $53.31\pm0.78$ 
                        & $66.82\pm0.17$ &$58.74\pm0.61$
                        & $72.92\pm0.30$&$63.57\pm0.49$
                        &$68.24\pm0.17$& $60.76\pm0.50$ 
                        & $63.18$ \\  
    \Xhline{3\arrayrulewidth}
    \end{tabular}
}
\end{center}
\caption{ Test Accuracy for symmetric noise (Equation~\ref{eqn:sym_noise}) removal from CIFAR10 and CIFAR100 datasets trained on VGG11 and  symmetric noise, asymmetric noise (Equation~\ref{eqn:asym_noise}) and hierarchical noise (Equation~\ref{eqn:real_noise}) removal ResNet18 architectures with noise strength $\eta=0.1$ and $\eta=0.25$. Note, the confusing class groups (or set $\mathcal{C}$) for hierarchical noise are cat-dog and truck-automobile for CIFAR10. For CIFAR100 we use Superclasses as the confusing groups. \\
\textbf{\textsuperscript{\textdagger}} - We select 1000 samples with low loss value from $\mathcal{D}_{Tr}$ as retain samples for our method. 
} \label{tab:cifar_additinal}
\end{table*}

\end{document}